\def\submissionblack{1} %
\documentclass[lettersize,journal]{IEEEtran}
\usepackage{amsmath,amsfonts}
\usepackage{algpseudocode}
\usepackage{algorithm}
\usepackage{array}
\usepackage[caption=false,font=normalsize,labelfont=sf,textfont=sf]{subfig}
\usepackage{textcomp}
\usepackage{stfloats}
\usepackage{url}
\usepackage{verbatim}
\usepackage{graphicx}
\usepackage{booktabs}
\usepackage{multirow}
\usepackage{ulem}
\usepackage{bm}
\usepackage{amsmath}
\usepackage{xcolor}
\usepackage{float}
\usepackage{capt-of} %
\usepackage[export]{adjustbox}
\usepackage{balance}
\usepackage{cite}

\newcommand{\qimg}[2][0.10\linewidth]{\raisebox{-.4\height}{\includegraphics[width=#1]{#2}}}
\newcommand{\imgw}{0.066\linewidth} %

\renewcommand{\qimg}[1]{\includegraphics[width=\imgw]{#1}}

\newcommand{\qimgm}[2]{%
  \IfFileExists{#1}{%
    \qimg{#1}%
  }{%
    \fbox{\parbox[c][\imgw][c]{\imgw}{\centering\scriptsize #2}}%
  }%
}

\usepackage{xcolor}

\definecolor{revR1col}{HTML}{1F6FB4}
\definecolor{revR2col}{HTML}{C0392B}
\definecolor{revR3col}{HTML}{27855A}
\definecolor{revAEcol}{HTML}{7F3FBF}

\newif\ifannot
\annottrue
\ifdefined\submissionblack\annotfalse\fi

\usepackage{eso-pic}
\AddToShipoutPictureFG*{\AtPageLowerLeft{\put(\LenToUnit{0.06\paperwidth},\LenToUnit{0.018\paperheight}){\parbox{0.88\paperwidth}{\fontsize{6.5}{7.5}\selectfont
\copyright~2026 IEEE. Personal use of this material is permitted. Permission from IEEE must be obtained for all other uses, in any current or future media, including reprinting/republishing this material for advertising or promotional purposes, creating new collective works, for resale or redistribution to servers or lists, or reuse of any copyrighted component of this work in other works. This is the authors' accepted version of an article published in IEEE Transactions on Visualization and Computer Graphics; the published version is available at \url{https://doi.org/10.1109/TVCG.2026.3738401}.}}}}
\begin{document}

\title{Hint-AUC: Deterministic Region-based Hint Generation for Line Art Colorization Evaluation}

\author{Koki Madono, Yuan Mingcheng, Edgar Simo-Serra
\thanks{Koki Madono, Yuan Mingcheng, Edgar Simo-Serra are with Waseda
University.}}
\markboth{IEEE Transactions on Visualization and Computer Graphics}%
{Madono \MakeLowercase{\textit{et al.}}: Hint-AUC: Deterministic Region-based Hint Generation for Line Art Colorization Evaluation}

\definecolor{sig25_revise}{rgb}{0.0, 0.0, 0.0}
\definecolor{tvcg26_revise}{rgb}{0.0, 0.0, 0.0}
\definecolor{tvcg26_1st_revise}{rgb}{0.0, 0.0, 0.0}
\ifdefined\submissionblack
\definecolor{tvcg26_2nd_revise}{rgb}{0.0, 0.0, 0.0}
\definecolor{tvcg26_2nd_layout}{rgb}{0.0, 0.0, 0.0}
\else
\definecolor{tvcg26_2nd_layout}{rgb}{0.0, 0.55, 0.0} %
\definecolor{tvcg26_2nd_revise}{rgb}{1.0, 0.0, 0.0}
\fi
\DeclareRobustCommand{\tvcgrev}[1]{\textcolor{tvcg26_1st_revise}{#1}}
\DeclareRobustCommand{\tvcgrevii}[1]{\textcolor{tvcg26_2nd_revise}{#1}}
\DeclareRobustCommand{\tvcglayout}[1]{\textcolor{tvcg26_2nd_layout}{#1}}

\newcommand{\vsd}[2]{\tvcgrev{$#1_{\,\pm #2}$}}
\newcommand{\vsdb}[2]{\tvcgrev{$\mathbf{#1}_{\,\pm #2}$}}
\newcommand{\vsdu}[2]{\tvcgrev{$\underline{#1}_{\,\pm #2}$}}

\newcommand{\xscr}{\tvcgrev{\text{scribble map}}} %

\maketitle

\begin{abstract}
Line art colorization is an important design application that has been extensively studied in recent years. 
Focusing on usability, most approaches have been developed for interactive use where colors are specified using user hints. 
However, collecting a large dataset of user hints for training is challenging. 
As a result, a variety of approaches have emerged for generating hints for both training and evaluation, 
such as color dots and scribbles.
Furthermore, the lack of a common evaluation framework, differences in hint generation implementations, 
and limited code availability reduce reproducibility.
\tvcgrev{For} these reasons, highly stochastic neural network-based algorithms have made it challenging to fairly evaluate and compare different approaches.
In this paper, we introduce a novel evaluation framework for line art colorization
algorithms that leverages deterministic region-based hints inspired by human drawing behavior.
Our framework enables in-depth, fair comparisons with existing colorization models.
Moreover, the same generated hints can be incorporated into training a colorization model
to significantly improve the quality of the colorization results\tvcgrev{, as shown in evaluations using both deterministically and randomly sampled hints}.
\tvcgrevii{Our evaluation is consistent with the overall human-preference results, and training with the generated hints improves colorization quality.}
\end{abstract}


\begin{figure*}[!t]
  \centering
  \includegraphics[width=0.74\textwidth,keepaspectratio]{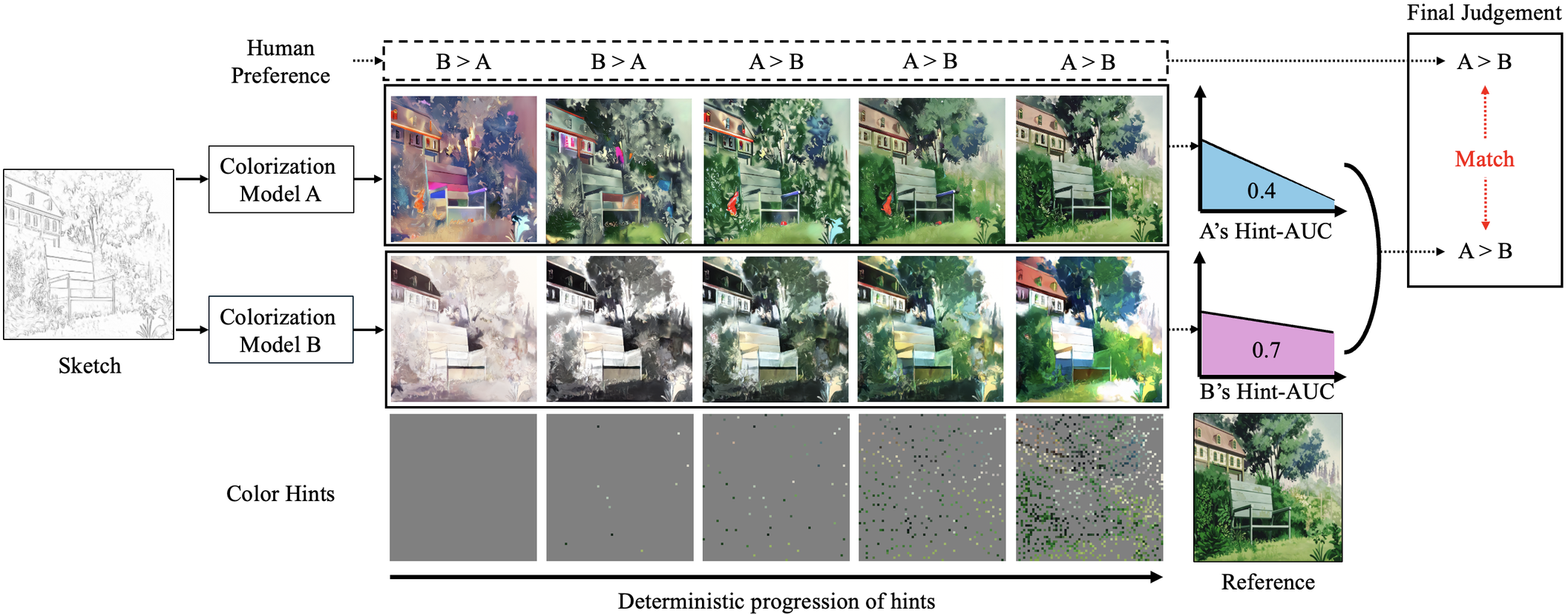}
  \caption{\textbf{Proposed Hint-AUC evaluation.} Recent interactive colorization methods employ varied evaluation schemes, making fair comparisons and model assessment difficult.
  In this work, we propose a new approach to evaluating interactive colorization, based on deterministic generation of hints and a holistic evaluation across interaction levels.
  \tvcgrevii{Hint-AUC is consistent with the overall human-preference results. The generated hints also improve model performance when used for training.}}
  \label{fig:teaser}
\end{figure*}

\section{Introduction}
\label{sec:intro}

Line art colorization plays an important role in illustration and design.
When done manually, an artist must use digital illustration tools (e.g., Photoshop) to fill in the color of the drawing.
\textcolor{sig25_revise}{Illustrations typically use a few base colors with additional complex tonal variations for lighting,
which makes them challenging for casual users to create from scratch.}
To speed up illustration, many approaches have focused on leveraging large
amounts of readily available illustration data to assist with the task, while enabling user control via hints such as scribbles and dots.
However, the difficulty of
obtaining the necessary large amounts of user hints as training data has
made approaches rely on heuristics to simulate user hints, typically as generated dots or scribbles.
Furthermore, the use of different techniques for data
augmentation, evaluation metrics, and hint types \tvcgrev{makes} it very
hard to interpret comparisons and results, due to a lack of a common evaluation
framework. 
\textcolor{tvcg26_1st_revise}{Such randomly generated dot or scribble hints 
have three drawbacks: exact results are not reproducible across runs or seeds unless the sampling seed is fixed and reported, 
the comparison depends on a single hint ratio at which scores 
are reported, and random placement does not reflect how users actually paint, such as coloring large regions first.}
A summary of recent approaches is shown in
Table~\ref{summary:recent-approach-evaluation}.

\begin{figure*}[t]
    \centering
    \includegraphics[keepaspectratio,width=0.80\linewidth]{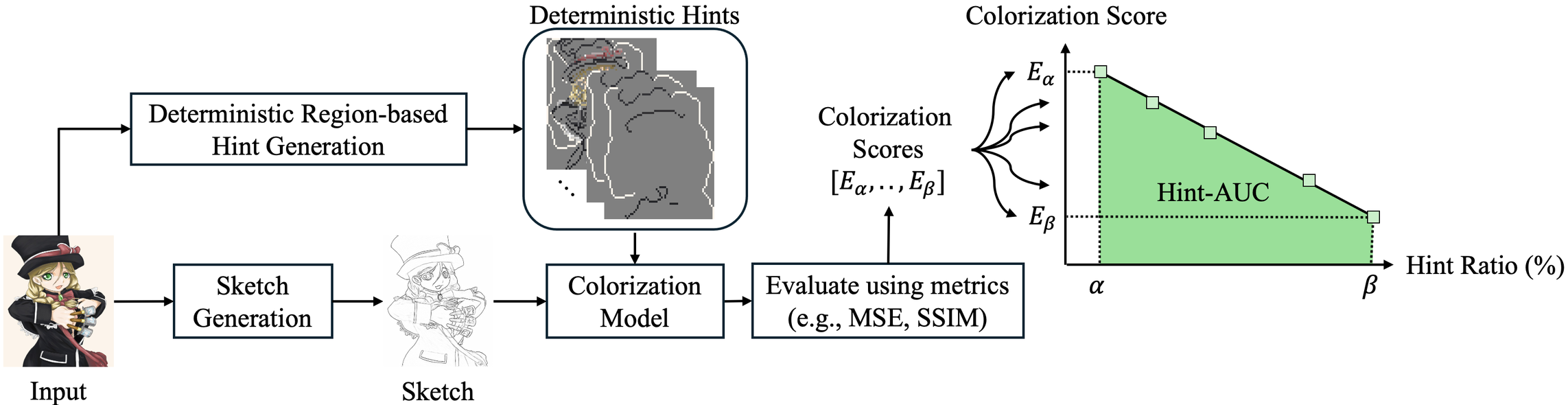}

    \caption{\textbf{Overview of the proposed Hint-AUC evaluation.}
    Deterministic color hints are generated progressively from the ground-truth image; performance 
    is then evaluated at multiple hint levels, 
    and finally we compute \textcolor{sig25_revise}{the area under} the resulting curve to obtain a robust metric for colorization model evaluation.}
\label{fig:overview-evaluation}
\end{figure*}

In this work, we present an evaluation framework for line art colorization with user-provided hints
that are both deterministic and in-depth. In particular, we propose \emph{Hint-AUC},
a new framework for user-guided line art
colorization evaluation, which consists of computing the Area Under the Curve
for common performance metrics with respect to the number of hints. 
\textcolor{tvcg26_1st_revise}{By \emph{stochastic evaluation} we mean 
a protocol whose hint positions are randomly sampled at test time, 
so that the reported scores vary across runs or seeds.}
To address the issue of stochastic evaluation, we develop a
deterministic region-based approach for hint generation, which, inspired by
human drawing behavior, focuses on providing hints for large important parts of
the image instead of minor details. As the number of provided hints increases,
the hints gradually go from coarse to fine, until all regions in the image are
\tvcgrevii{selected}. By considering the entire Area Under the Curve, our
proposed Hint-AUC is able to more accurately capture the full behavior of
user-hint-guided art colorization approaches.
An overview is shown in \textcolor{sig25_revise}{Fig.}~\ref{fig:teaser}.

We show how the same framework can also be used during training to
significantly improve the colorization performance, and propose a
conditional diffusion-based model called \emph{DetFill} for line art colorization.
Our model outperforms the compared line art colorization approaches across diverse evaluation settings and
demonstrates more consistent performance as the number of hints increases.

In summary, our main contributions \textcolor{sig25_revise}{are} as follows:
\begin{itemize}
\setlength{\parskip}{0cm}
\setlength{\itemsep}{0cm}
   \item A framework, named \emph{Hint-AUC}, for fair evaluation and comparison of hint-based colorization approaches based on deterministic hint generation.
\item We show that the same framework \tvcgrev{can also be} used to improve colorization results, and propose \emph{DetFill}, a \tvcgrev{conditional diffusion-based line art colorization model that demonstrates the benefits of the proposed evaluation protocol}.
   \item In-depth evaluation of our proposed framework including a perceptual user study.
\end{itemize}

\section{Related Work}
\label{sec:related-word}

\noindent\textbf{User-Guided Line Art Colorization.}
User-guided line art colorization has been developed over the years using
several types of user-provided guidance in order to control the colorization.
\textcolor{sig25_revise}{
Many methods have achieved impressive results
on colorization of natural photographs~\cite{zabari2023diffusing,cong2024automatic}.
However, \textcolor{tvcg26_revise}{line art} colorization poses additional challenges: line drawings consist solely of sparse binary contours,
offering neither grayscale gradients nor texture cues for semantic inference.
Moreover, automatic \textcolor{tvcg26_revise}{line art} generation algorithms~\cite{winnemoller2012xdog,sketchkeras,simo2018mastering}
often produce imperfect or noisy strokes, further complicating reliable color propagation.
In light of these difficulties, several recent studies
have begun to address colorization specifically within the \textcolor{tvcg26_revise}{line art} domain.
}

In terms of guidance, \tvcgrev{reference-image-based}
approaches~\cite{Yan_2025_WACV, yan2025colorizediffusionv2enhancingreferencebased, Zhang_2025_ICCV, dai2024learning,xing2024tooncrafter,huang2024lvcd,10.1145/3343031.3351041,10.2312:egs.20201023,9143503,Li_2022_CVPR,li2022eliminating,10182273,liu2025manganinja,meng2024anidoc}
use global and local information for colorization, \tvcgrev{and text-, color-, image-, and multimodal-guided
approaches~\cite{cong2024automatic,10.1145/3664647.3681356,yin2024coloredit,liang2024control} incorporate
complementary guidance signals into the colorization process.} \textcolor{tvcg26_revise}{However, these approaches 
can be less practical for workflows that require direct, localized color control.}

A more common approach is to use dots or scribbles drawn directly on the line art as user hints.
This provides more intuitive control over the
colorization process, but the number and locations of the user hints can have a significant impact on the generated images, and this varies across approaches.
Early approaches were based on heuristics~\cite{qu-2006-manga,Sykora09-EG},
and these approaches were later extended to the learning-based approaches~\cite{Sangkloy_2017_CVPR,10.1016/j.neucom.2018.05.045,xia-2018-invertible,9515572,fang2025vangoghunifiedmultimodaldiffusionbased,zhuang2025cobraefficientlineart}.
In this work, we target single-stage scribble and dot-based
approaches~\cite{10.1145/3240508.3240661,Filling2021zhang,Hati2019PaintsTorchAU,CarrilloCVPRW2023},
and focus on how to more accurately compare and evaluate different approaches
without depending heavily on the heuristic process and testing parameters
chosen for easier and more reliable comparisons.

\begin{table*}[t]
   \caption{\textbf{Recent line art colorization approaches.} We summarize recent data-driven user-guided single-stage line art colorization approaches, focusing on the type of hints and evaluation.
   $\bigtriangleup$ indicates the dataset is partially unavailable.}
  \label{summary:recent-approach-evaluation} %
  \centering %
  \begin{tabular}{cccccccc}
     Approach & Year &  Code & Train Data  & Test Data & Hint Type & Evaluation  \\
        \toprule
        Paint Chainer~\cite{PaintChainer} & 2017&  $\checkmark$ & & &N/A &N/A \\
        Tandem~\cite{frans2017outline} & 2017 & & & & Blurred Image & Only visualization\\
        Petalica Paint~\cite{Petalica} & 2018 & & & & N/A   &N/A \\
        Auto Painter~\cite{10.1016/j.neucom.2018.05.045} & 2018 & $\checkmark$ & & & Dots &  User study\\
        AlacGAN~\cite{10.1145/3240508.3240661} & 2018 &   $\checkmark$  & & &  Scribbles/Dots & FID\\
        PaintsTorch~\cite{Hati2019PaintsTorchAU} & 2019 &  $\checkmark$  & & & Scribbles/Dots & FID, Standard Deviation\\
        Dual Color Colorization~\cite{9515572} & 2021& & &  &  Scribbles &MOS, FID\\
        SFM~\cite{Filling2021zhang} & 2021 & & $\checkmark$ & &Scribbles&MAE, User study\\
        Diffusart~\cite{CarrilloCVPRW2023} & 2023&  $\checkmark$ & $\checkmark$ & $\checkmark$ & Scribbles &SSIM, FID, LPIPS\\
        
        \textcolor{tvcg26_revise}{Cobra}~\cite{zhuang2025cobraefficientlineart} &  2025 & $\checkmark$ & &  & Reference Image Pool & CLIP, PSNR, SSIM, AS~\cite{schuhmann2022laion}\\
        \textcolor{tvcg26_revise}{Manga Ninja}~\cite{liu2025manganinja} &  2025 & $\checkmark$ & $\checkmark$ & $\checkmark$ & Image from Movie & DINO, CLIP, PSNR, SSIM, LPIPS\\

        \textcolor{tvcg26_revise}{MagicColor}~\cite{Zhang_2025_ICCV} &  2025 & $\checkmark$ & $\bigtriangleup$ & $\bigtriangleup$& Reference Image and Segmentation & FID, PSNR, SSIM, LPIPS\\
        \textcolor{tvcg26_revise}{ColDiffv1}~\cite{Yan_2025_WACV} &  2025 & $\checkmark$ & $\checkmark$ & $\checkmark$ & Reference image &FID\\
        \textcolor{tvcg26_revise}{ColDiffv2}~\cite{yan2025colorizediffusionv2enhancingreferencebased} &  2025 & $\checkmark$ & $\checkmark$ & $\checkmark$ & Reference image &SSIM, FID, LPIPS\\
        \bottomrule
  \end{tabular}
\end{table*}

\noindent\textbf{Colorization Evaluation.}
In user-guided line art colorization, quantitative metrics and user studies are
commonly used for evaluation. In particular, perceptual loss-based
metrics such as FID~\cite{NIPS2017_8a1d6947} and
LPIPS~\cite{zhang2018perceptual} have been widely used as evaluation metrics, with more recent variants such as DINO~\cite{oquab2023dinov2} and DreamSim~\cite{fu2023dreamsim}.
Pixel-level
metrics such as MSE~\cite{4775883}, PSNR~\cite{huynh2008scope}, and
SSIM~\cite{1284395} have also been adapted to the task. 
However, when evaluating with generated color hints, 
\textcolor{tvcg26_1st_revise}{these scores are typically 
reported under inconsistent and often stochastic hint-sampling 
protocols---e.g., a single, arbitrarily chosen hint ratio with randomly placed hints---which makes 
fair comparison across methods difficult.}

User studies are often utilized in the evaluation and are a high-quality way
to evaluate colorization techniques. SFM~\cite{Filling2021zhang}
measures interaction time with user attribute and auxiliary color table,
Fang~\cite{9412756} uses an absolute rating method, and ArchGAN~\cite{9240535}
uses a Mean Opinion Score (MOS).
Quantitative metrics typically consider only the generated results, and do not account for user guidance information.
A user study is limited in sample diversity and requires
knowledge of colorization for users. 
\textcolor{sig25_revise}{Furthermore, user studies can be costly and final outcomes are highly dependent on the quality of both study design and participants.}
Our work proposes a methodological and consistent evaluation that considers the hint
information.

\noindent\textbf{Hint Generation Algorithms.}
The most common approach to creating color hints for line art colorization is to randomly sample
the number and location of hints from the uniform
distribution~\cite{Hati2019PaintsTorchAU, 10.1145/3240508.3240661,
CarrilloCVPRW2023}. The scribbles can then be flattened to be a single color.
However, the random sampling approach often tends to converge to the dominant color
information on the image and can lead to an unstable evaluation, and in
the worst case, adding more hints can actually be detrimental to performance.
In our approach, we deterministically generate hints using region information
of \tvcgrev{the} target colorized line art.
We generate hints deterministically and apply them in a fixed sequence,
enabling a more comprehensive and reproducible evaluation of colorization algorithms.

\section{Formative Study}
\label{sec:formative_study}

\textcolor{sig25_revise}{
We performed a formative study, in which participants used digital painting software with an iterative workflow to obtain insights on human drawing behaviors.
From their interactions with line art colorization, we analyzed how they apply color,
which inspired our perceptual hint evaluation metric to more accurately reflect those human behaviors.
}

\subsection{Participants}

\textcolor{sig25_revise}{
    We recruited 16 users for the formative study.
    All users have experience using computers, and seven users have at least two years of experience using illustration software.
    We obtained consent from all participants to use their drawing process data and survey responses,
    while ensuring their personal information remained protected.
}
\subsection{Experimental Setting}
\textcolor{sig25_revise}{We used Krita\footnote{https://krita.org/}, an open-source painting application with
a rich set of brushes and editing tools, as our experiment interface.
In order to capture all the user interactions, we implemented a custom plugin
that saves every user action, while also saving the entire canvas
periodically.}

\textcolor{sig25_revise}{For stimuli, we prepared four line art images in total with two tasks:
(1) A free-form task using two line arts (case 1, 2) where no colorization example was provided to allow unrestricted colorization.
(2) A reference-based task using \tvcgrev{two} different line arts (case 3, 4) where the user can refer to a full-color exemplar during line art colorization.
}

\subsection{Procedure}

\textcolor{sig25_revise}{Each participant ran through a one-hour session structured as follows:
\begin{enumerate}
   \item Explanation of the experiment with a practice session (10 min): We gave participants a brief introduction to Krita’s interface and brush tools, then had them colorize a simple line drawing to familiarize themselves with the workflow.
   \item Colorization Tasks (45 min total): We provided each participant with two line art images, counterbalancing both the task order and the image types across users to cover all possible sequences.
      Each participant spent at least 15 minutes on each task, then \tvcgrev{was} given five additional minutes to refine their colorization until they were satisfied with the result.
   \item Survey (5 min total): We provided a survey about participants' experience with computers and colorization. Furthermore, we asked about their strategies of colorization.
\end{enumerate}
By comparing performance on free-form versus reference-based tasks,
we can examine how having an exemplar influences participants’ color-selection behaviors and workflow.
To enable direct comparisons of participants’ performance and their experience with
each task, we adopted a within-subject design.
}

\begin{figure}[ht]
        \centering
        \begin{tabular}{c@{\hspace{0.01\linewidth}}c@{\hspace{0.01\linewidth}}c@{\hspace{0.01\linewidth}}c}    
        \vspace{1mm}
        Case 1 &
        \raisebox{-.4\height}{\includegraphics[ width = .27\linewidth]{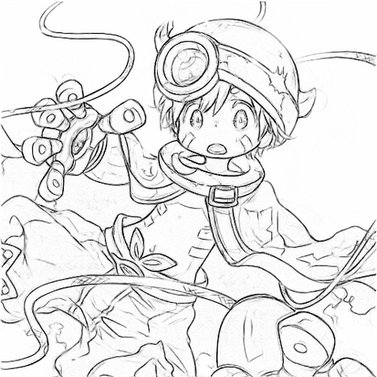}} &
        \raisebox{-.4\height}{\includegraphics[ width = .27\linewidth]{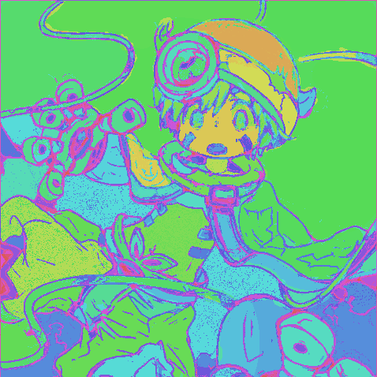}} &
        \raisebox{-.4\height}{\includegraphics[ width = .27\linewidth]{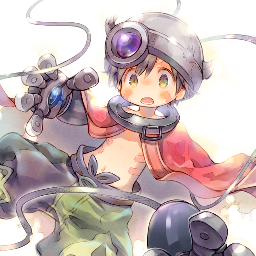}} \\        
        
        Case 2 &
        \raisebox{-.4\height}{\includegraphics[ width = .27\linewidth]{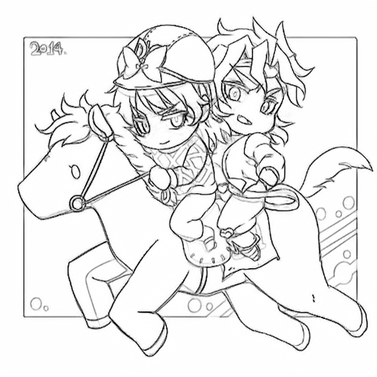}} &
        \raisebox{-.4\height}{\includegraphics[ width = .27\linewidth]{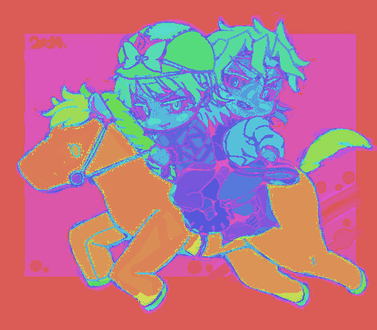}} &
        \raisebox{-.4\height}{\includegraphics[ width = .27\linewidth]{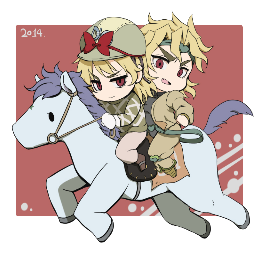}} \\        
        Case 3 &
        \raisebox{-.4\height}{\includegraphics[ width = .27\linewidth]{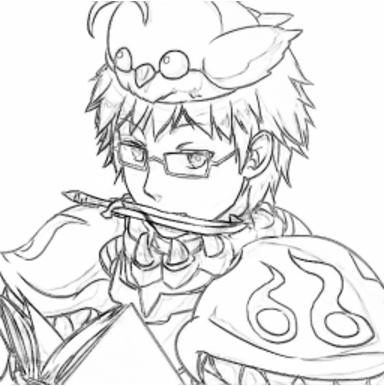}} &
        \raisebox{-.4\height}{\includegraphics[ width = .27\linewidth]{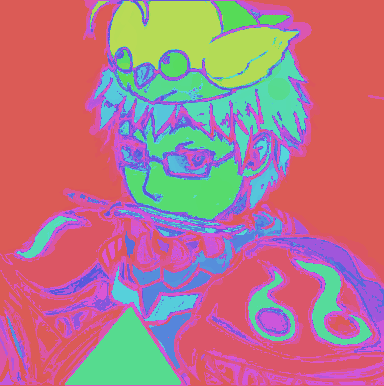}} &
        \raisebox{-.4\height}{\includegraphics[ width = .27\linewidth]{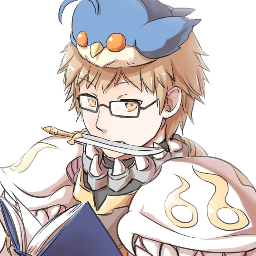}} \\        

        Case 4 &
        \raisebox{-.4\height}{\includegraphics[ width = .27\linewidth]{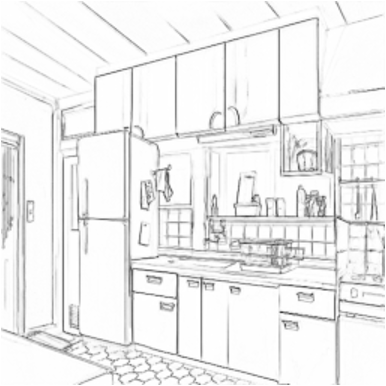}} &
        \raisebox{-.4\height}{\includegraphics[ width = .27\linewidth]{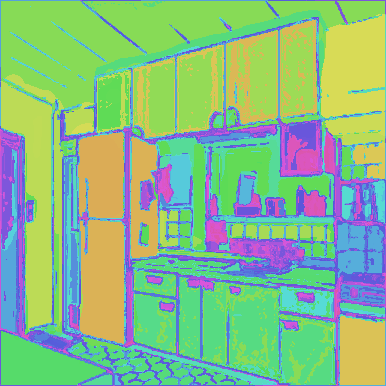}} &
        \raisebox{-.4\height}{\includegraphics[ width = .27\linewidth]{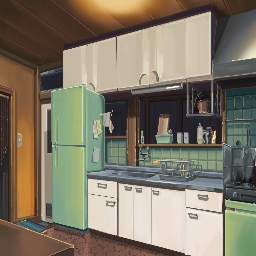}} \\        

        & & \raisebox{-.4\height}{\includegraphics[ width = .27\linewidth]{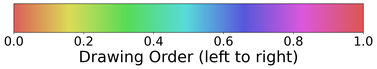}} & \\        
        
        & {Line Art} & {\tvcgrev{Progress Heatmap}} &  {GT}  \\ 
    \end{tabular}
    \caption{\textbf{User Colorization Process Analysis:} We generate a colorization progress heatmap, 
    where regions painted earlier appear in red and regions painted later appear in blue.}        
    \label{fig:user_colorization_progress_analysis}
\end{figure}
            
\begin{figure}[ht]
    \centering
    \begin{tabular}{c@{\hspace{0.01\linewidth}}c@{\hspace{0.01\linewidth}}c@{\hspace{0.01\linewidth}}c}    
    \vspace{1mm}
    
    \raisebox{-.4\height}{\includegraphics[ width = .34\linewidth]{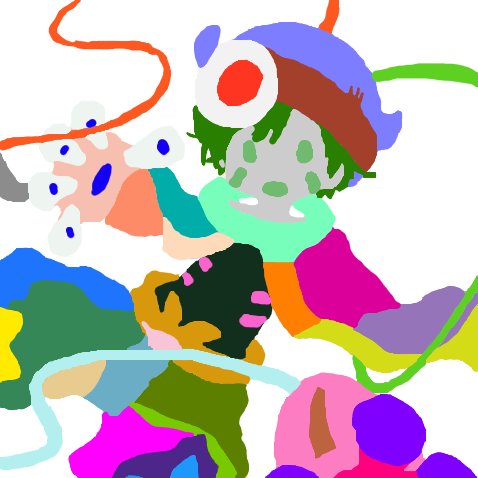}} &
    \raisebox{-.4\height}{\includegraphics[ width = .61\linewidth]{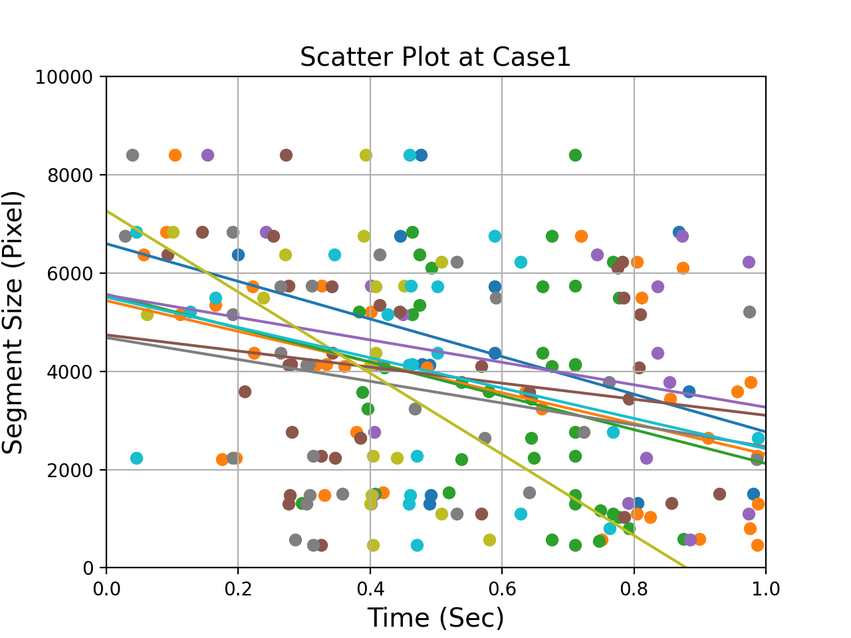}} \\        

    \raisebox{-.4\height}{\includegraphics[ width = .34\linewidth]{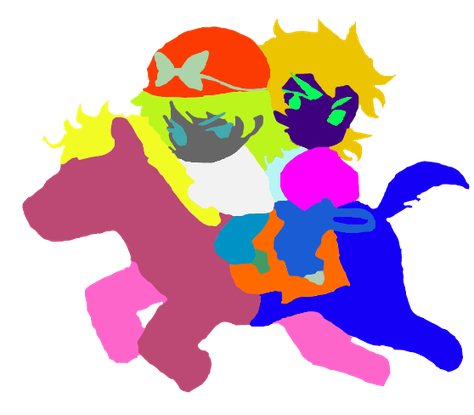}} &
    \raisebox{-.4\height}{\includegraphics[ width = .61\linewidth]{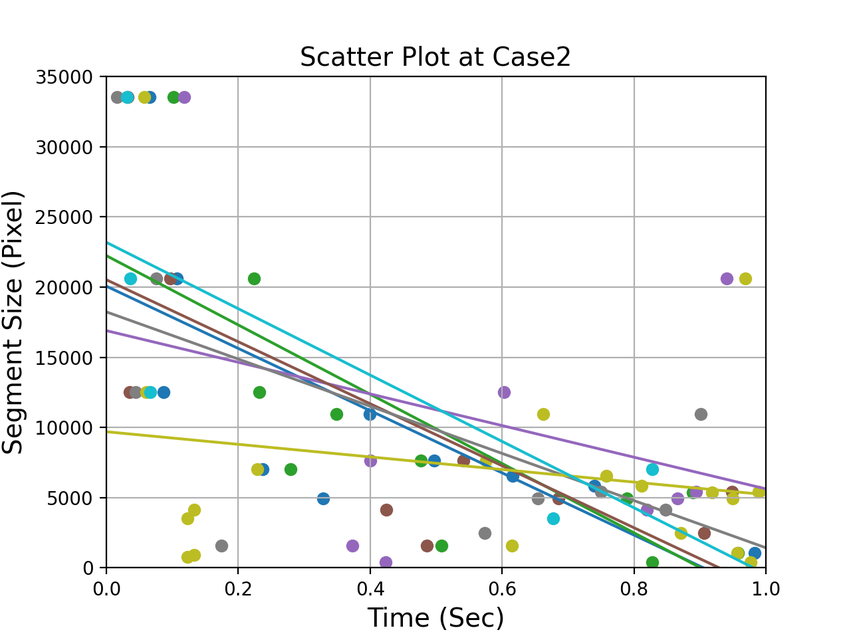}} \\        

    \raisebox{-.4\height}{\includegraphics[ width = .34\linewidth]{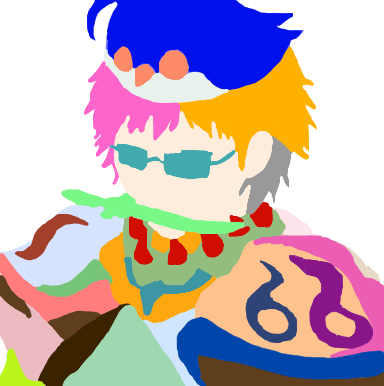}} &
    \raisebox{-.4\height}{\includegraphics[ width = .61\linewidth]{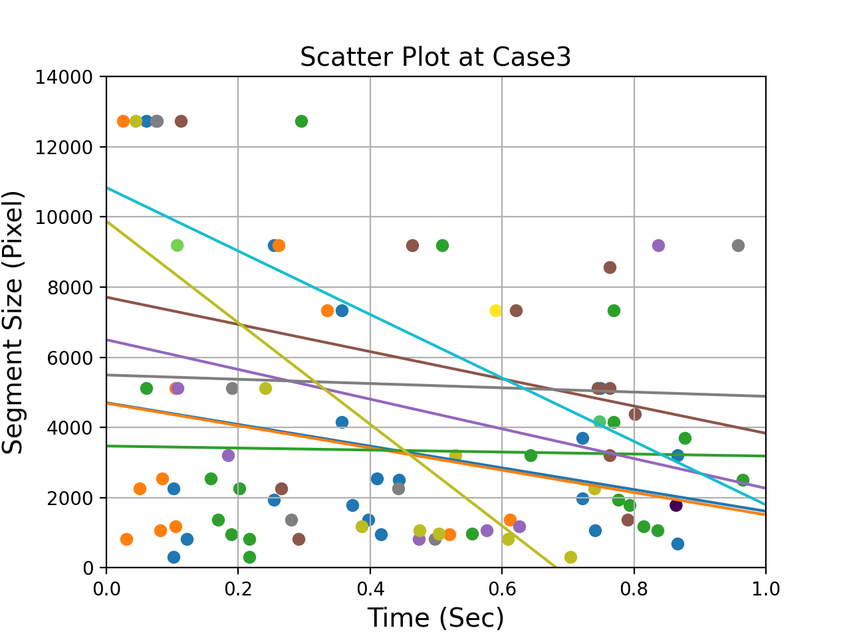}} \\        

    \raisebox{-.4\height}{\includegraphics[ width = .34\linewidth]{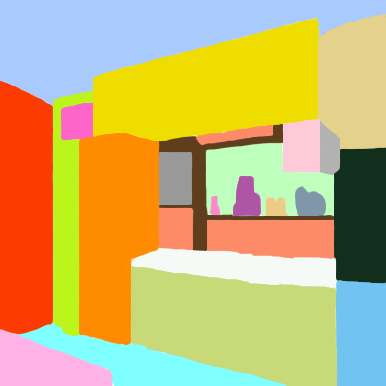}} &
    \raisebox{-.4\height}{\includegraphics[ width = .61\linewidth]{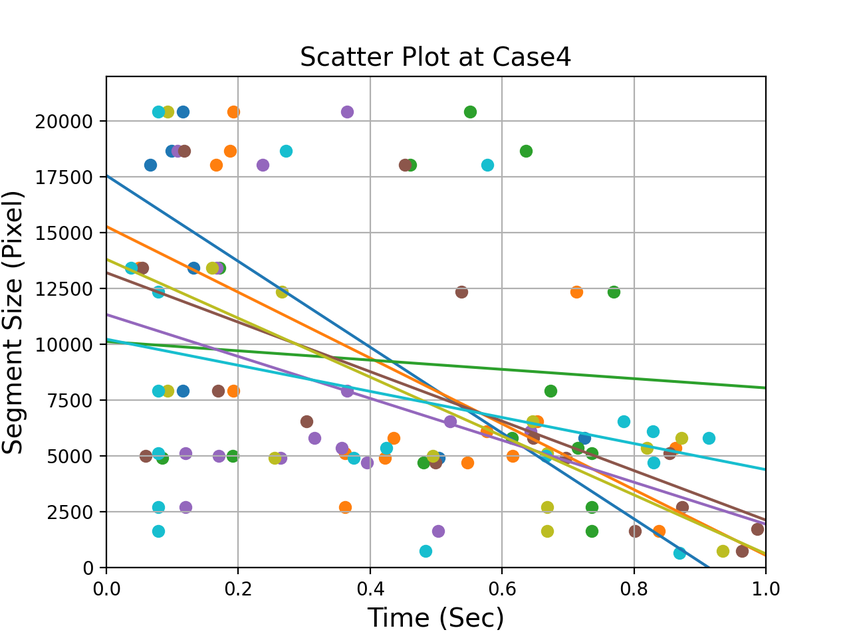}} \\        

    {Segmentation} & {Scatter Plot}  \\ 
\end{tabular}
\caption{
    \textbf{Segment Filling Analysis:} Using the cleaned segmentation map, we track the sequence in which each user fills regions. 
    We plot each user’s data in a distinct color and fit a linear regression relating segment size to fill order. 
    A negative slope therefore reflects a tendency to color larger regions earlier and smaller regions later.
}        
\label{fig:persegment_filing_analysis}
\end{figure}
        
\begin{figure*}[ht]
    \centering
    \begin{tabular}
    {c@{\hspace{0.01\linewidth}}c@{\hspace{0.01\linewidth}}c@{\hspace{0.01\linewidth}}c@{\hspace{0.01\linewidth}}c@{\hspace{0.01\linewidth}}c@{\hspace{0.01\linewidth}}c@{\hspace{0.01\linewidth}}c@{\hspace{0.01\linewidth}}}
    
    \vspace{1mm}
    Beginner1 & 
    \raisebox{-.4\height}{\includegraphics[ width = .12\linewidth]{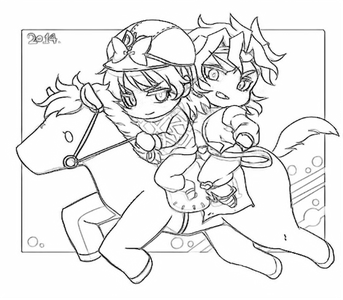}} &
    \raisebox{-.4\height}{\includegraphics[ width = .12\linewidth]{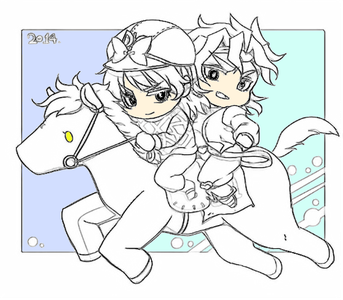}} &
    \raisebox{-.4\height}{\includegraphics[ width = .12\linewidth]{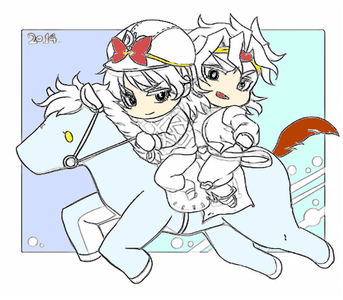}} &
    \raisebox{-.4\height}{\includegraphics[ width = .12\linewidth]{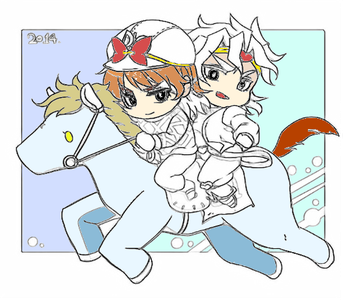}} &
    \raisebox{-.4\height}{\includegraphics[ width = .12\linewidth]{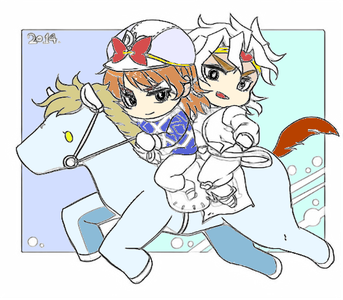}} &
    \raisebox{-.4\height}{\includegraphics[ width = .12\linewidth]{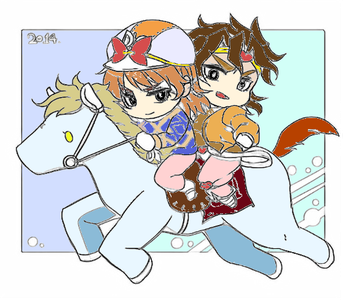}} &
    \raisebox{-.4\height}{\includegraphics[ width = .12\linewidth]{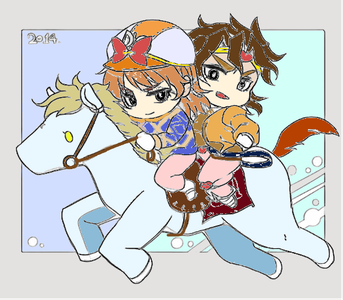}}\\
    Beginner2 &  
    \raisebox{-.4\height}{\includegraphics[ width = .12\linewidth]{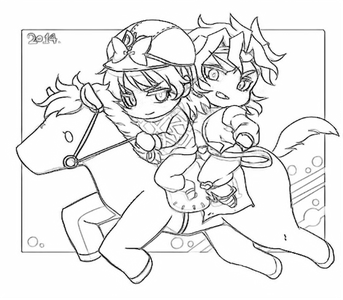}} &
    \raisebox{-.4\height}{\includegraphics[ width = .12\linewidth]{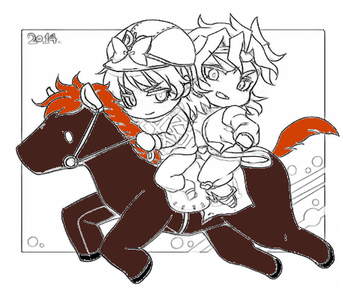}} &
    \raisebox{-.4\height}{\includegraphics[ width = .12\linewidth]{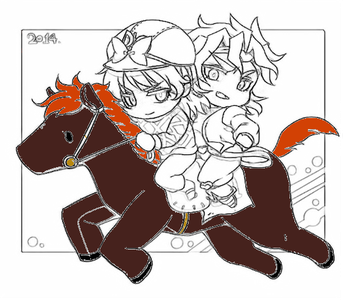}} &
    \raisebox{-.4\height}{\includegraphics[ width = .12\linewidth]{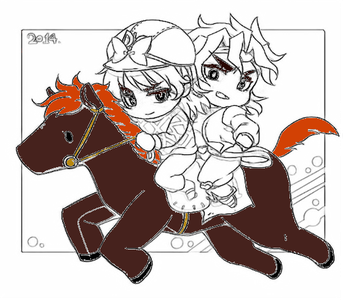}} &
    \raisebox{-.4\height}{\includegraphics[ width = .12\linewidth]{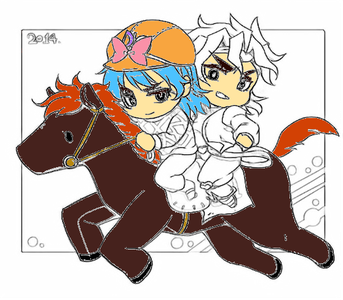}} &
    \raisebox{-.4\height}{\includegraphics[ width = .12\linewidth]{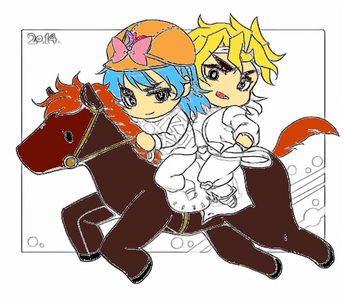}} &
    \raisebox{-.4\height}{\includegraphics[ width = .12\linewidth]{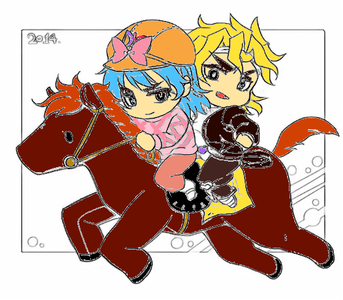}} \\
    Expert & 
    \raisebox{-.4\height}{\includegraphics[ width = .12\linewidth]{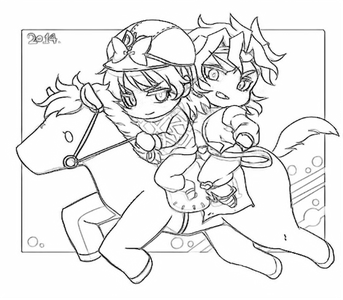}} &
    \raisebox{-.4\height}{\includegraphics[ width = .12\linewidth]{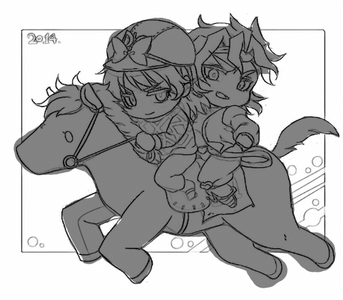}} &
    \raisebox{-.4\height}{\includegraphics[ width = .12\linewidth]{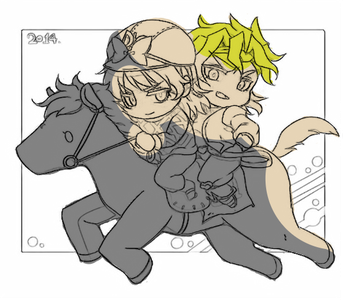}} &
    \raisebox{-.4\height}{\includegraphics[ width = .12\linewidth]{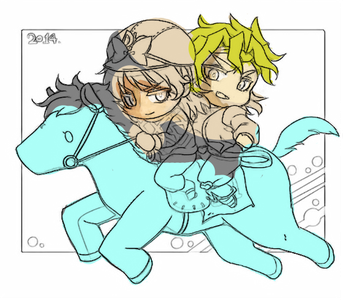}} &
    \raisebox{-.4\height}{\includegraphics[ width = .12\linewidth]{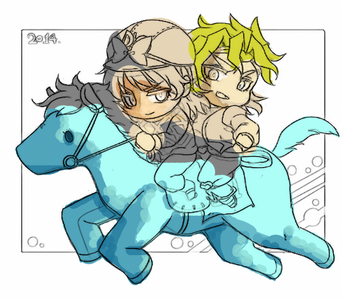}} &
    \raisebox{-.4\height}{\includegraphics[ width = .12\linewidth]{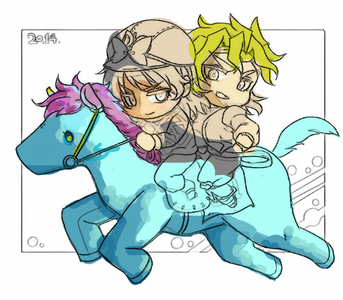}} &
    \raisebox{-.4\height}{\includegraphics[ width = .12\linewidth]{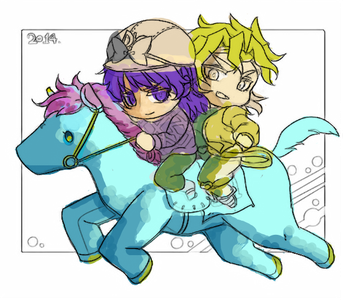}} \\

    \vspace{1mm}
    & {0 min} & {3 min} & {6 min} & {9 min} & {12 min} & {15 min} & {Final} \\

    \end{tabular}
    \caption{\textbf{User Colorization Process with a non-reference image:\ }%
    Beginners have basic computer skills and the expert has two years of experience creating illustrations. 
    All of them have their own final outcome image and different colorization behaviors.}
    \label{fig:exp1_visualization_colorization_unsup}
    \end{figure*}

\subsection{Findings from Users}
\label{ssec:finding_from_user}
\textcolor{sig25_revise}{
We illustrate the user colorization process using a heatmap in Fig.~\ref{fig:user_colorization_progress_analysis}
and visualize the free-form colorization processes of several users in Fig.~\ref{fig:exp1_visualization_colorization_unsup}.
From these results, we distilled three primary characteristics of human drawing behaviors, summarized as follows:}

\textcolor{sig25_revise}{\textbf{Users tend to fill large regions first:}
We examine users’ filling behavior using the prepared segmentation (Fig.~\ref{fig:persegment_filing_analysis}).
Specifically, we record, for each user, the normalized time at which they have filled at least
$N\%$ of a segment, with $N=70\%$ chosen to balance sample size against the decision threshold for ``completion''. We then compute, on a per-user basis,
a linear regression of segment area versus normalized fill time, and can see
that all show a downward tendency, i.e., the users fill larger regions first.
To confirm this trend statistically, we perform one-sided tests of the null hypothesis that each slope is zero against the alternative that it is negative.
\tvcgrev{At the individual level, among the eight users who completed each case, significant negative slopes ($p<0.05$) are observed in 7 users for case 1, 4 for case 2, 2 for case 3, and 4 for case 4.}
In the aggregated analysis pooling all users, each case also yields $p<0.05$, confirming the overall tendency.
As illustrated in Fig.~\ref{fig:exp1_visualization_colorization_unsup},
some participants use bucket tools to apply base colors to multiple regions at once.
Although a handful of small segments are occasionally filled early (and vice versa),
the dominant pattern is that users prioritize coloring larger areas at the start.
}

\tvcgrev{\textbf{A top-first tendency appears in some cases:}}
\textcolor{sig25_revise}{Fig.~\ref{fig:user_colorization_progress_analysis} visualizes users’ painting procedures with heatmaps:
red regions denote areas painted early, and blue regions denote areas painted later.
In cases 1 and 3, early-stage painting (red) is concentrated at the top of the images.
}
\tvcgrev{The tendency is clearest in Case~1, weaker in Cases~3 and~4, and absent in Case~2.}
\tvcgrev{To quantify this, for each (user, case) we computed the Spearman
correlation $\rho$ between paint-event time and the y-centroid of newly
painted pixels; positive $\rho$ indicates top-first painting. Across the
16 analyzed users (each case was completed by eight users), the pattern is clearest in Case~1 (7/8 users with $\rho>0$, 3/8 significant
at one-sided $p<0.05$, Wilcoxon signed-rank $p=0.039$); it is weaker in
Case~3 (5/8 positive, mean $\rho=+0.12$) and Case~4 (5/8 positive, mean
$\rho=+0.05$), and absent in Case~2 (0/8 significant, mean $\rho=-0.09$).
These numbers quantitatively support the summary above. This tendency is a descriptive observation
about user behavior only; it is not used in our hint-generation algorithm,
which orders regions solely by descending area.}

\textcolor{sig25_revise}{\textbf{Each user has distinct drawing behaviors:}
When examining Fig.~\ref{fig:exp1_visualization_colorization_unsup},
we observe that users adopt distinct colorizing behaviors.
Beginners typically fill areas one at a time, but the sequence in which
they choose regions varies widely. Experts, by contrast, usually begin by blocking in
large regions and then refine subtle color transitions to achieve more natural tones.
These findings indicate that no single behavior can capture the full diversity of
users’ colorization approaches.}

\subsection{Considerations for Evaluation}
\label{ssec:design_hint_eval}
\textcolor{sig25_revise}{Based on the formative study,
we summarize two key requirements that an ideal
line art evaluation metric should satisfy from the user’s perspective.}

\begin{enumerate}
   \item \textcolor{sig25_revise}{\textbf{R1: Evaluate images sequentially:}
    As shown in Fig.~\ref{fig:user_colorization_progress_analysis},
    users gradually refine their colorizations over time. When assessing a user’s coloring ability,
    it is therefore essential to consider not only the final result but also the sequence of intermediate steps.
    Similarly, to fully evaluate a colorization model, we need to examine its outputs at multiple stages of the coloring process.}
    \item \textcolor{sig25_revise}{\textbf{R2: Evaluate large regions first:}
    Since users typically fill in large regions before attending to finer details,
    it is best to begin colorization model evaluation with those larger areas.
    Accordingly, we must generate hints deterministically from a segmentation map
    that includes both large and small regions, giving priority to hints in the larger segments.
    }
\end{enumerate}

\textcolor{sig25_revise}{
    As discussed in Section~\ref{ssec:finding_from_user},
    incorporating additional user characteristics into our hint evaluation metric is
    challenging and risks overfitting to specific users.
    To address these findings, we introduce the Hint-AUC metric in the next section.
}

\section{Proposed Line Art Colorization Evaluation}

Our approach focuses on creating a robust and deterministic measure for
evaluating line art colorization algorithms.
\textcolor{tvcg26_revise}{
Inspired by our findings in the formative study, our evaluation is designed to mimic the user colorization process by focusing on the large regions first.
Furthermore, by jointly considering many different numbers of hints, we can reduce the dependency on the hint generation parameters during evaluation.}
In particular, we propose
\textcolor{sig25_revise}{Hint \textcolor{tvcg26_revise}{Area Under the Curve}, abbreviated as \emph{Hint-AUC}},
which considers a wide range of color hints as part of the
performance measure by computing the \textcolor{tvcg26_revise}{Area Under the metric–hint Curve} of performance
metrics with an increasing number of color hints that are computed
deterministically.
Fig.~\ref{fig:overview-evaluation} illustrates an overview of the proposed evaluation.

\subsection{Deterministic Region-based Hint Generation}
\label{ssec:det_hint_generation}

\begin{figure*}[t]
    \centering
    \includegraphics[keepaspectratio,width=0.80\linewidth]{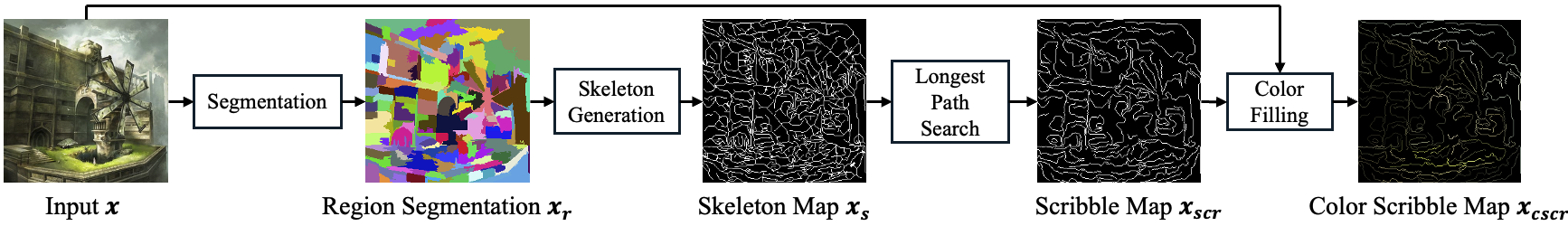}

\caption{\textbf{Deterministic region-based hint generation.}
First, we perform region segmentation and use this to obtain a skeleton image.
Afterwards, we compute scribbles by sampling the longest path from each skeleton image.
Finally, we compute the scribble color by using the average color of the region.
We use region size to progressively add hints which can then be used to compute the Hint-AUC metric.
}
\label{fig:proposed-method-hint-generation}
\end{figure*}

We generate deterministic region-based color hints in three steps: segmentation, skeleton extraction, and color filling.
\tvcgrevii{For evaluation we use fixed stored hint and label maps; regenerating them from the source images is not bit-exact because two preprocessing steps are unseeded (supplementary Sec.~I).}
\textcolor{sig25_revise}{Fig.~\ref{fig:proposed-method-hint-generation} illustrates
an overview of our deterministic region-based hint generation.}

\begin{figure*}[t]
    \centering
    \includegraphics[keepaspectratio,width=0.80\linewidth]{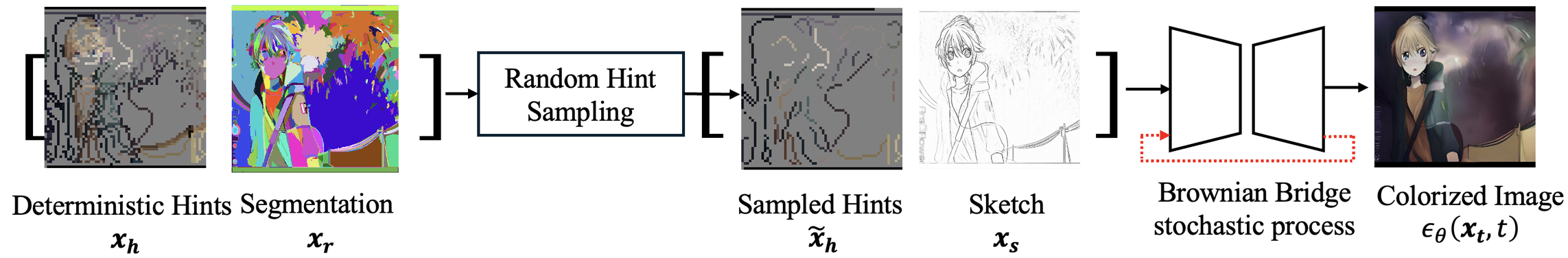}
    \caption{\textbf{Overview of proposed colorization model DetFill.}
    We use the Hint-AUC framework to sample sets of hints during training, and use them to train a BBDM-based model\textcolor{sig25_revise}{~\cite{li2023bbdm}} to colorize images.}
\label{fig:overview-model-arch}
\end{figure*}

\vspace{0.15em}
\noindent\textbf{Region Segmentation.}
The segmentation task \textcolor{sig25_revise}{used to obtain image regions for colorization} is defined as follows:
\begin{equation}
   \bm{x}_{r} = G_{r}(\bm{x})\;,
\end{equation}
\noindent where $\bm{x}$ denotes the input image, $\bm{x}_{r}$ denotes the
region segmentation maps, and $G_{r}$ denotes the segmentation algorithm
that splits the image into a set of
non-overlapping regions, each representing a single dominant color.
\tvcgrev{Since models are trained against a fixed $G_r$, the test-time
segmenter should come from the same region-respecting family (see the
supplementary material).}

\vspace{0.15em}
\noindent\textbf{Skeleton Generation.}
For each region $i$,
\textcolor{sig25_revise}{to obtain a color hint candidate,}
we independently compute the region skeleton map by using a
morphological operator-based thinning algorithm~\cite{10.1145/357994.358023}.
Specifically, we select the $i$-th binary region mask $\bm{x}_{r}^{i}$ from the segmentation map $\bm{x}_{r}$ and use it as
input for skeleton generation as follows:
\begin{equation}
   \bm{x}_{s}^{i} = f(\bm{x}_{r}^{i}) \;,
\end{equation}
\noindent where $\bm{x}_{s}^{i}$ denotes the $i$-th skeleton map, and $f$ denotes
the thinning algorithm~\cite{10.1145/357994.358023}.

\vspace{0.15em}
\textcolor{sig25_revise}{\noindent\textbf{Longest Path Search.}}
Given the skeleton map $\bm{x}_{s}^{i}$,
we estimate the corresponding scribble and dot hints.

For scribble color hints, we treat each skeleton region as a graph and compute
the maximum path on the graph~\cite{2015MNRAS.452.3435K}, which we then render
as a raster image:
\begin{equation}
   \bm{x}_{\mathrm{scr}}^{i} = g(h(\bm{x}_{s}^{i}, \delta))\;,
\end{equation}
\noindent where $\bm{x}_{\mathrm{scr}}^{i}$ denotes \textcolor{tvcg26_revise}{a binary scribble map}, $h$ denotes a dilation function
to adjust the thickness of the skeleton, $g$ denotes a longest path search function, and $\delta$ denotes the
thickness parameter \textcolor{tvcg26_1st_revise}{(a $3{\times}3$ kernel in our implementation).
The resulting longest path is intersected with the region mask $\bm{x}_{r}^{i}$,
yielding a one-pixel-wide scribble inside the region}.

\textcolor{sig25_revise}{\noindent\textbf{Color Filling.}}
We apply the color to each scribble \textcolor{sig25_revise}{mask} as follows:
\begin{equation}
   \bm{x}_{\mathrm{cscr}}^{i} = \zeta(\bm{x}_{\mathrm{scr}}^{i}, \bm{x})\;,
   \label{eq:col_avg}
\end{equation}
\noindent where $\bm{x}_{\mathrm{cscr}}^{i}$ denotes a color scribble map, $\zeta$ denotes a color averaging operation
\textcolor{sig25_revise}{which consists of simply taking the average color value of the region
for each scribble using both \tvcgrev{the} \xscr~$\bm{x}_{\mathrm{scr}}^{i}$ and input image~$\bm{x}$.}

\tvcgrevii{From the binary scribble map $\bm{x}_{\mathrm{scr}}$ we also derive the binary dot map $\bm{x}_{\mathrm{dot}}$ and the color dot map $\bm{x}_{\mathrm{cdot}}$. We place each dot at the mean row and column coordinates of the longest-path pixels, truncated to integers; the coordinates are computed before intersecting the path with the region mask. The dot is not projected onto its source region and can therefore fall outside a nonconvex region. The dot color is the mean color of the region that contains the dot pixel. The supplementary material reports an estimate of out-of-region placement based on the stored scribble masks.}
\textcolor{tvcg26_1st_revise}{We deliberately assign a single averaged color
per region: hints serve as sparse color control signals rather than as
supervision for shading. Encoding lighting, shadows, or color gradients into
hints is itself a challenging open problem, as such effects are global
functions of light direction, materials, and artistic style rather than
per-region properties.}
\tvcgrevii{Each region produces at most one stroke. Scribbles are clipped to their source regions, and each hint pixel has a single color.}

\vspace{0.15em}
\subsection{Region-based Hint Sampling}
\textcolor{sig25_revise}{Although we can generate color hints deterministically,
it is equally important to order the hints in a deterministic manner to compute the Hint-AUC.
Building on the insight from Section~\ref{ssec:finding_from_user}
that users color large regions first, we order regions by size, from largest to smallest.}
\tvcgrevii{At hint ratio $\alpha$, we select the $k=\lfloor\alpha n\rfloor$ largest regions ($\alpha n$ rounded down). Equal-area regions are ordered by the sorting routine of the implementation, and this order depends on the library version, as described in supplementary Sec.~I. For $\alpha>0$, an image with $n<1/\alpha$ receives no hints (one of the 3,000 test images at $\alpha{=}1\%$). Here, $\alpha$ is the fraction of regions selected, not a fixed hint-pixel or information budget. For an image with $n$ regions, $k$ approximates the number of hint interactions. With scribbles, selecting larger regions first may yield longer strokes and greater pixel coverage than area-independent selection at the same $\alpha$.}
\textcolor{sig25_revise}{Unlike conventional colorization methods that randomly sample hints~\cite{Hati2019PaintsTorchAU,CarrilloCVPRW2023},
colors within each segmented region tend to be homogeneous. Therefore, when sampling hints from a region,
the average scribble color computed using Eq.~\eqref{eq:col_avg} is expected to exhibit low variance.}

\textcolor{sig25_revise}{\subsection{Hint-AUC}}
\label{ssec:hau-c}
Hint-AUC simultaneously considers various hints to provide a per-image evaluation.
\textcolor{sig25_revise}{Specifically, we use our deterministic hints and vary the hint ratio $\alpha$
to generate \textcolor{tvcg26_revise}{the metric–hint curve} by computing the metric scores (e.g., MSE, PSNR, or LPIPS) for each hint ratio $\alpha$.
We then calculate the \textcolor{tvcg26_revise}{area under the metric–hint curve},
commonly referred to as the AUC.}
\textcolor{sig25_revise}{Hint-AUC is computed with the following equation:}
\begin{equation}
   \mathrm{HAUC} = \frac{1}{2} \sum_{i=1}^{N-1} \left(s(\hat{\bm{y}}_{\alpha_{i+1}}, \bm{y}) + s(\hat{\bm{y}}_{\alpha_{i}}, \bm{y})\right)\,\left(\alpha_{i+1} - \alpha_i \right) \;,
\end{equation}
\noindent where $\alpha_i$ denotes the $i$-th hint ratio,
$\hat{\bm{y}}_{\alpha_{i}}$ denotes the colorization output obtained when providing hints at ratio $\alpha_i$,
$s$ denotes an evaluation metric function~\textcolor{sig25_revise}{(e.g., MSE, PSNR)},
$\bm{y}$ denotes the ground-truth image, and $N$ is the number of steps to use when approximating the AUC\@.
\textcolor{tvcg26_1st_revise}{We compute the HAUC independently for each test image 
and report its unweighted mean over the entire test set as the final Hint-AUC score, 
so that each (metric, method, hint type) yields a single value. When several line-art sources 
are available, we additionally report the mean~$\pm$~sample standard deviation across 
the three sketch sources.}

\section{DetFill Colorization Model}

We propose \emph{DetFill}, a conditional diffusion model for line art colorization
inspired by BBDM~\cite{li2023bbdm}.
\textcolor{sig25_revise}{Fig.~\ref{fig:overview-model-arch}
shows an overview of DetFill.}
\textcolor{sig25_revise}{
Other diffusion-based line art colorization models~\cite{CarrilloCVPRW2023}
utilize latent features from hints via a cross-attention module~\cite{vaswani2017attention}
within the diffusion process~\cite{ho2020denoising} for colorization.
However, this makes the resulting colorization hard to control and thus harder 
to achieve the desired colorization in the final output.
}
In contrast, by using a Brownian Bridge stochastic process, our approach is able to directly
convert the sketch and \textcolor{sig25_revise}{hints} into a colorized image.

\textcolor{sig25_revise}{\subsection{DetFill Architecture}}

DetFill uses a \mbox{U-Net} backbone, inspired by Guided-diffusion~\cite{dhariwal2021diffusion}.
Specifically, the model uses a stack of residual layers with convolutional operations and skip connections.
\tvcgrev{Additionally, the model includes a single self-attention block at the \tvcgrevii{$64\times64$} bottleneck;
since the sketch and hints are concatenated with the diffusion state at the input,
the bottleneck features already incorporate the conditioning information when self-attention is applied.}

\textcolor{sig25_revise}{In contrast to similar architectures such as Diffusart~\cite{CarrilloCVPRW2023},
DetFill also incorporates the hints into the input features, leading to more accurate colorization results.
Another difference is that the model
uses \tvcgrev{96 (scribble) / 64 (dot) base channels in the \mbox{U-Net}} to be able to fully denoise the input images.
For full details of the architecture, please refer to the supplementary material.
}

\textcolor{sig25_revise}{\subsection{Training Details}}

During training, we (i) randomly sample deterministic hints and (ii) concatenate the sketch and color hints as model inputs.
Specifically, we sample the deterministic hints during training using the following equation.
\begin{equation}
   \bm{x}_{\hat{h}} = \bm{M}_{\mathcal{I}} \odot \bm{x}_{h}\;,
\end{equation}
\noindent where $\bm{x}_{\hat{h}}$ denotes the sampled hints,
$\bm{x}_{h}$ denotes the deterministic hints of the input image $\bm{x}$,
$\bm{S}$ denotes the region label map of $\bm{x}$ (with $n$ unique labels),
$\mathcal{I} \subseteq \{1,\dots,n\}$ is a set of selected labels with $|\mathcal{I}| = p$ (sampled uniformly at random),
and $\bm{M}_{\mathcal{I}}$ is the corresponding binary mask (1 for pixels whose label in $\bm{S}$ is in $\mathcal{I}$, 0 otherwise).
\tvcgrevii{For each training example, we independently draw $u\sim\mathcal{U}[0,1)$, randomly permute the $n$ region labels, and retain the first $p=\lfloor nu\rfloor$. Thus, $p$ is uniform on $\{0,\ldots,n-1\}$. Training includes zero-hint inputs but excludes $p=n$, which is used only at evaluation ($\alpha=1$).}
\textcolor{sig25_revise}{
The goal of randomly sampling deterministic hints during training is to ensure robust colorization inference,
regardless of the number of hints or their locations.
}

\tvcgrev{The Brownian Bridge stochastic process starts from the ground-truth state
$\bm{x}_{0}$ at $t{=}0$ and ends at the conditional input state $\bm{x}_{T}$ at $t{=}T$.
The diffused state $\bm{x}_{t}$ and the training objective $\tilde{\bm{y}}_{t}$ at timestep $t$ are}
\begin{equation}
   \bm{x}_{t} = \left(1 - m_{t}\right)\bm{x}_{0} + m_{t}\,\bm{x}_{T} + \sqrt{\delta_{t}}\,\bm{\epsilon},
   \qquad
   \tilde{\bm{y}}_{t} = \bm{x}_{t} - \bm{x}_{0},
\end{equation}
\tvcgrev{\noindent where $m_{t}$ \tvcgrevii{takes $T$ equally spaced values from $0.001$ to $0.999$ (so that $\delta_{t}>0$ at both ends)}, $\delta_{t} = 2s\left(m_{t} - m_{t}^{2}\right)$ is the
Brownian-Bridge variance with scale $s$ ($s{=}1$ in our experiments), and
$\bm{\epsilon} \sim \mathcal{N}\left(\bm{0}, \bm{I}\right)$ is standard Gaussian noise.}

\tvcgrev{We sample the timestep $t$ uniformly from $\{0,\ldots,T{-}1\}$ and
compute the loss as follows.}
\begin{equation}
   \mathcal{L} = \left\| \tilde{\bm{y}}_{t} - \epsilon_{\theta}\left(\bm{x}_{t}, t\right) \right\|_{1} + \lambda_{P}P\left(\tilde{\bm{y}}_{t}, \epsilon_{\theta}\left(\bm{x}_{t}, t\right)\right)  \;,
\end{equation}
\noindent where \tvcgrev{$\tilde{\bm{y}}_{t}$ denotes the training objective at timestep $t$},
$\epsilon_{\theta}$ denotes the diffusion model, $\lambda_{P}$ is a weighting hyperparameter \tvcgrev{($\lambda_{P}{=}0.1$; $P$ is computed on the RGB channels)}, and \tvcgrev{$\bm{x}_{T}$} denotes concatenated inputs of sketch and randomly sampled deterministic hints $\bm{x}_{\hat{h}}$,
and binary mask of hint location $\bm{x}_{\hat{hb}}$\tvcgrev{, while $\bm{x}_{0}$ is the corresponding ground-truth state}.
As a primary loss, we follow \cite{saharia2022palette} and use an \textcolor{sig25_revise}{$L_{1}$ loss}.
We additionally use LPIPS~\cite{zhang2018perceptual} as the perceptual loss $P$, as it is lightweight yet effective in constraining the clarity of the output.
\tvcgrev{Since the objective directly encodes the clean state
($\hat{\bm{x}}_{0} = \bm{x}_{t} - \epsilon_{\theta}\left(\bm{x}_{t}, t\right)$),
DetFill effectively learns the mapping from the sketch-and-hint condition to the target colorization.}

\textcolor{sig25_revise}{\subsection{Inference Details}}

\tvcgrev{During inference, we concatenate the input sketch, deterministic hints, and the corresponding mask to form the conditional state $\bm{x}_{T}$,
and perform $T^{\prime}$-step denoising ($T^{\prime}{=}200$) over a decreasing subsequence
$\tvcgrevii{T-1} = t_{0} > t_{1} > \cdots > t_{\tvcgrevii{T^{\prime}-1}} = 0$ of the original $T{=}1000$ timesteps,
accelerating the sampling with a non-Markovian process~\cite{song2020denoising}.
Starting from $\hat{\bm{y}}_{t_{0}} = \bm{x}_{T}$, each step computes:
}
\begin{equation}
\begin{split}
   \hat{\bm{y}}_{t_{k+1}}
   &= \frac{\delta_{t_{k+1}}\left(1 - m_{t_{k}}\right)}{\delta_{t_{k}}\left(1 - m_{t_{k+1}}\right)}\hat{\bm{y}}_{t_{k}}
      + \frac{\left(1 - m_{t_{k}}\right)\delta_{t_{k}, t_{k+1}}}{\delta_{t_{k}}}\hat{\bm{x}}_{0} \\
      &\quad + \left(m_{t_{k+1}} - \frac{m_{t_{k}}\left(1-m_{t_{k}}\right)\delta_{t_{k+1}}}{\left(1-m_{t_{k+1}}\right)\delta_{t_{k}}}\right)\bm{x}_{T}
       + \sigma_{k} \bm{\epsilon}
\end{split}
\end{equation}
\tvcgrev{\noindent where $\hat{\bm{x}}_{0} = \hat{\bm{y}}_{t_{k}} - \epsilon_{\theta}\left(\hat{\bm{y}}_{t_{k}}, t_{k}\right)$ denotes the reconstructed clean state,
$m_{t}$ and $\delta_{t}$ follow the training schedule \tvcgrevii{(defined on the original $T$ timesteps)},
$\delta_{t_{k}, t_{k+1}}$ denotes the variance gap computed as
$\left(\delta_{t_{k}} - \delta_{t_{k+1}}\frac{\left(1-m_{t_{k}}\right)^2}{\left(1-m_{t_{k+1}}\right)^2}\right)$,
$\bm{\epsilon} \sim \mathcal{N}\left(\bm{0},\bm{I}\right)$ is standard Gaussian noise, and the noise scale is
$\sigma_{k} = \sqrt{\frac{\delta_{t_{k}, t_{k+1}}\,\delta_{t_{k+1}}}{\delta_{t_{k}}}}$.
\tvcgrevii{At the final step ($t_{T^{\prime}-1}{=}0$), we return $\hat{\bm{x}}_{0}$ directly.} The final colorization is given by the RGB channels of $\hat{\bm{y}}_{0}$.}

\textcolor{sig25_revise}{This inference algorithm can reduce the number of timesteps $T^{\prime}$ compared to the number of timesteps during the training phase
$T$ and contribute to reducing the inference time.}

\providecommand{\qimg}[1]{}

\begin{figure*}[t]
\centering
\setlength{\tabcolsep}{6.5pt} %
\renewcommand{\arraystretch}{1.0} %

\scriptsize
\begin{tabular}{c c c c c c c c c}

\raisebox{4\height}{$1\%$} &
\qimgm{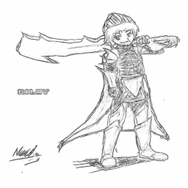}{sketch B} &
\qimgm{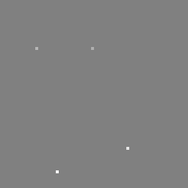}{hint B} &
\qimgm{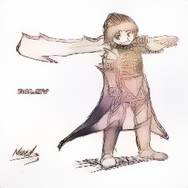}{PT B} &
\qimgm{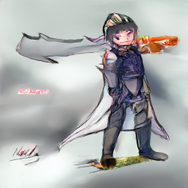}{Diffusart B} &
\qimgm{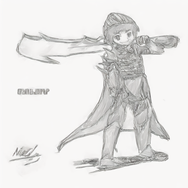}{CDv1 B} &
\qimgm{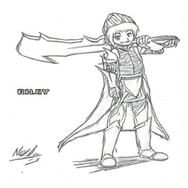}{CDv2 B} &
\qimgm{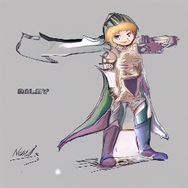}{DetFill B} &
\qimgm{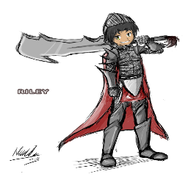}{GT B} \\
\raisebox{4\height}{$10\%$} &
\qimgm{figs/imgs_sig25/exp1_rev/2875_sketch.png}{sketch B} &
\qimgm{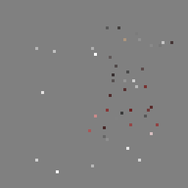}{hint B} &
\qimgm{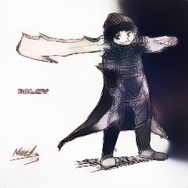}{PT B} &
\qimgm{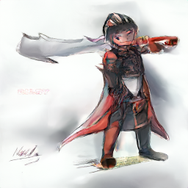}{Diffusart B} &
\qimgm{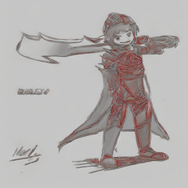}{CDv1 B} &
\qimgm{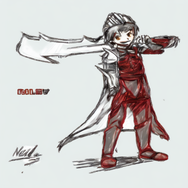}{CDv2 B} &
\qimgm{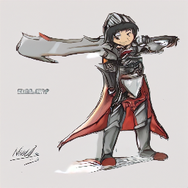}{DetFill B} &
\qimgm{figs/imgs_sig25/exp1_rev/proposed/748016.image_groundtruth.png}{GT B} \\

\raisebox{4\height}{$50\%$} &
\qimgm{figs/imgs_sig25/exp1_rev/2875_sketch.png}{sketch B} &
\qimgm{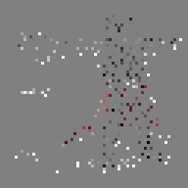}{hint B} &
\qimgm{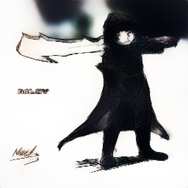}{PT B} &
\qimgm{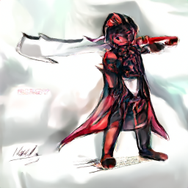}{Diffusart B} &
\qimgm{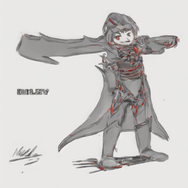}{CDv1 B} &
\qimgm{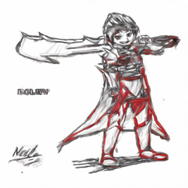}{CDv2 B} &
\qimgm{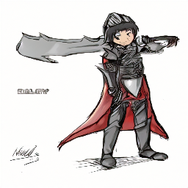}{DetFill B} &
\qimgm{figs/imgs_sig25/exp1_rev/proposed/748016.image_groundtruth.png}{GT B} \\

\raisebox{4\height}{$100\%$} &
\qimgm{figs/imgs_sig25/exp1_rev/2875_sketch.png}{sketch B} &
\qimgm{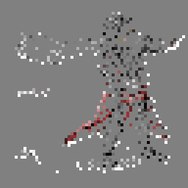}{hint B} &
\qimgm{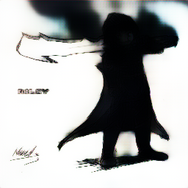}{PT B} &
\qimgm{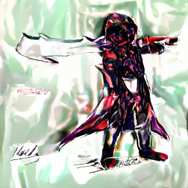}{Diffusart B} &
\qimgm{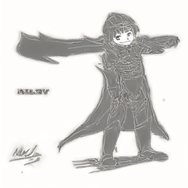}{CDv1 B} &
\qimgm{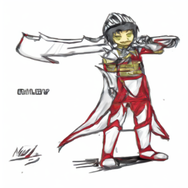}{CDv2 B} &
\qimgm{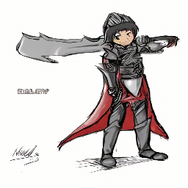}{DetFill B} &
\qimgm{figs/imgs_sig25/exp1_rev/proposed/748016.image_groundtruth.png}{GT B} \\

\raisebox{4\height}{$1\%$} &
\qimgm{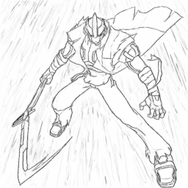}{sketch A} &
\qimgm{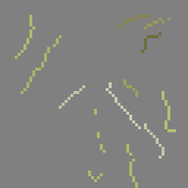}{hint A} &
\qimgm{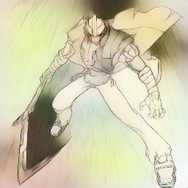}{PT A} &
\qimgm{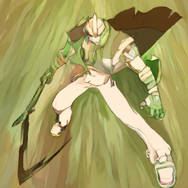}{Diffusart A} &
\qimgm{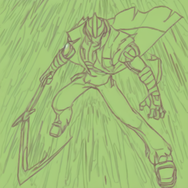}{CDv1 A} &
\qimgm{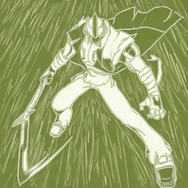}{CDv2 A} &
\qimgm{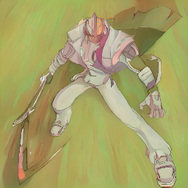}{DetFill A} &
\qimgm{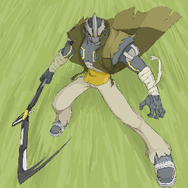}{GT A} \\

\raisebox{4\height}{$10\%$} &
\qimgm{figs/imgs_sig25/exp1_rev/1429_sketch.png}{sketch A} &
\qimgm{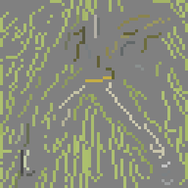}{hint A} &
\qimgm{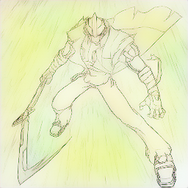}{PT A} &
\qimgm{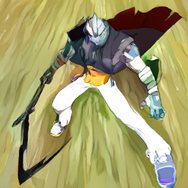}{Diffusart A} &
\qimgm{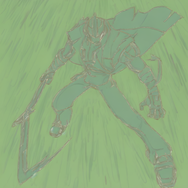}{CDv1 A} &
\qimgm{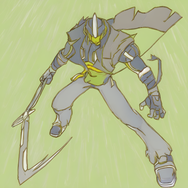}{CDv2 A} &
\qimgm{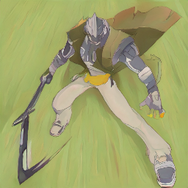}{DetFill A} &
\qimgm{figs/imgs_sig25/exp1_rev/proposed/514016.image_groundtruth.png}{GT A} \\

\raisebox{4\height}{$50\%$} &
\qimgm{figs/imgs_sig25/exp1_rev/1429_sketch.png}{sketch A} &
\qimgm{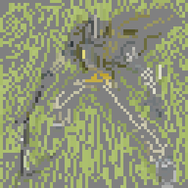}{hint A} &
\qimgm{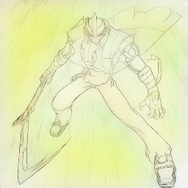}{PT A} &
\qimgm{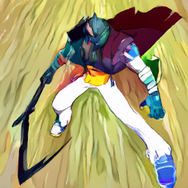}{Diffusart A} &
\qimgm{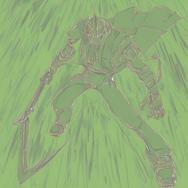}{CDv1 A} &
\qimgm{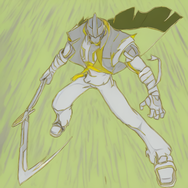}{CDv2 A} &
\qimgm{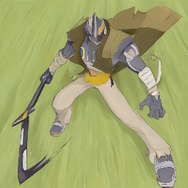}{DetFill A} &
\qimgm{figs/imgs_sig25/exp1_rev/proposed/514016.image_groundtruth.png}{GT A} \\

\raisebox{4\height}{$100\%$} &
\qimgm{figs/imgs_sig25/exp1_rev/1429_sketch.png}{sketch A} &
\qimgm{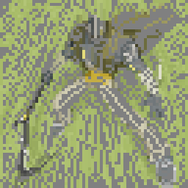}{hint A} &
\qimgm{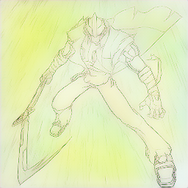}{PT A} &
\qimgm{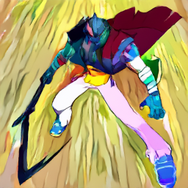}{Diffusart A} &
\qimgm{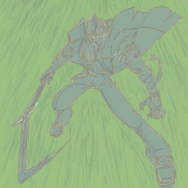}{CDv1 A} &
\qimgm{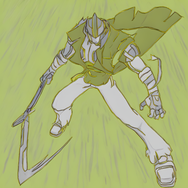}{CDv2 A} &
\qimgm{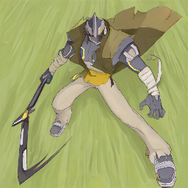}{DetFill A} &
\qimgm{figs/imgs_sig25/exp1_rev/proposed/514016.image_groundtruth.png}{GT A} \\

\textbf{$\alpha$} &
\textbf{Sketch} &
\textbf{Hints} &
\shortstack{\textbf{PaintsTorch}} &
\shortstack{\textbf{Diffusart}} &
\shortstack{\textbf{ColDiffv1}} &
\shortstack{\textbf{ColDiffv2}} &
\textbf{DetFill} &
\textbf{GT} \\

\end{tabular}

\caption{\textcolor{tvcg26_1st_revise}{\textbf{Qualitative comparison under different hint ratios.}
For each $\alpha$ (rows) we show two examples (A, B). All models are trained
under the common 200-epoch protocol; the authors' original ColorizeDiffusion
fine-tuning schedule is evaluated quantitatively in the supplementary
material.}}
\label{fig:exp1_visualization_colorization}
\end{figure*}

\begin{table*}[t]
  \caption{\textbf{\tvcglayout{Hint-AUC on Danbooru2021 with size-ordered region-based hints.}}
   \tvcglayout{Values are the mean $\pm$ sample standard deviation across three sketch sources (XDoG, sketch simplification, and SketchKeras; $n=3$)\tvcgrevii{, not across random seeds}. The label \emph{200ep} denotes ColorizeDiffusion fine-tuned from the official weights under our 200-epoch protocol. The original 7-epoch schedule and the Diffusart retraining ablation are evaluated in the supplementary material.}
   }%
  \label{tbl:exp1-result-hintauc-deterministic}
   \vspace{-2mm}
  \centering
  \footnotesize
  \setlength{\tabcolsep}{4pt}
  \begin{tabular}{cc ccccccc}
    \toprule
      \multirow{2}{*}{Hint} & \multirow{2}{*}{Model} &   \multicolumn{7}{c}{Hint-AUC} \\
     & &    MSE ($\downarrow$) & PSNR ($\uparrow$) & SSIM ($\uparrow$)  & LPIPS ($\downarrow$) & OpenCLIP ($\uparrow$) & DINO ($\uparrow$)  & DreamSim ($\downarrow$)   \\
\midrule
\multirow{5}{*}{Dot}
& \multirow{1}{*}{PaintsTorch~\cite{Hati2019PaintsTorchAU}} &   0.079\textcolor{tvcg26_1st_revise}{$\pm$0.002} & 11.012\textcolor{tvcg26_1st_revise}{$\pm$0.06} & 0.480\textcolor{tvcg26_1st_revise}{$\pm$0.027} & 0.484\textcolor{tvcg26_1st_revise}{$\pm$0.024} & 0.820\textcolor{tvcg26_1st_revise}{$\pm$0.017} & 0.849\textcolor{tvcg26_1st_revise}{$\pm$0.011} & 0.464\textcolor{tvcg26_1st_revise}{$\pm$0.018}  \\
& \multirow{1}{*}{Diffusart~\cite{CarrilloCVPRW2023}} &  0.093\textcolor{tvcg26_1st_revise}{$\pm$0.004} & 10.398\textcolor{tvcg26_1st_revise}{$\pm$0.10} & 0.470\textcolor{tvcg26_1st_revise}{$\pm$0.022} & 0.396\textcolor{tvcg26_1st_revise}{$\pm$0.008} & 0.877\textcolor{tvcg26_1st_revise}{$\pm$0.003} & 0.866\textcolor{tvcg26_1st_revise}{$\pm$0.002} & 0.323\textcolor{tvcg26_1st_revise}{$\pm$0.002}  \\
   & \textcolor{tvcg26_revise}{\multirow{1}{*}{ColDiffv1 \textcolor{tvcg26_1st_revise}{(200ep)}~\cite{Yan_2025_WACV}}} &  \textcolor{tvcg26_revise}{\underline{0.054\textcolor{tvcg26_1st_revise}{$\pm$0.007}}} & \textcolor{tvcg26_revise}{13.623\textcolor{tvcg26_1st_revise}{$\pm$0.60}} & \textcolor{tvcg26_revise}{\underline{0.537\textcolor{tvcg26_1st_revise}{$\pm$0.056}}}  &
   \textcolor{tvcg26_revise}{0.375\textcolor{tvcg26_1st_revise}{$\pm$0.058}} & \textcolor{tvcg26_revise}{0.873\textcolor{tvcg26_1st_revise}{$\pm$0.036}}  & \textcolor{tvcg26_revise}{0.902\textcolor{tvcg26_1st_revise}{$\pm$0.034}}  &  \textcolor{tvcg26_revise}{0.350\textcolor{tvcg26_1st_revise}{$\pm$0.093}} \\
   & \textcolor{tvcg26_revise}{\multirow{1}{*}{ColDiffv2 \textcolor{tvcg26_1st_revise}{(200ep)}~\cite{yan2025colorizediffusionv2enhancingreferencebased}}} &  \textcolor{tvcg26_revise}{0.059\textcolor{tvcg26_1st_revise}{$\pm$0.011}} & \textcolor{tvcg26_revise}{\underline{13.825\textcolor{tvcg26_1st_revise}{$\pm$0.76}}}  & \textcolor{tvcg26_revise}{0.523\textcolor{tvcg26_1st_revise}{$\pm$0.039}}  & \textcolor{tvcg26_revise}{\underline{0.335\textcolor{tvcg26_1st_revise}{$\pm$0.031}}}
    & \textcolor{tvcg26_revise}{\underline{0.898\textcolor{tvcg26_1st_revise}{$\pm$0.018}}} & \textcolor{tvcg26_revise}{\underline{0.909\textcolor{tvcg26_1st_revise}{$\pm$0.020}}} & \textcolor{tvcg26_revise}{\underline{0.253\textcolor{tvcg26_1st_revise}{$\pm$0.041}}}\\
   & \multirow{1}{*}{DetFill (Ours)} &   \textbf{0.021\textcolor{tvcg26_1st_revise}{$\pm$0.004}} & \textbf{17.321\textcolor{tvcg26_1st_revise}{$\pm$0.91}} & \textbf{0.675\textcolor{tvcg26_1st_revise}{$\pm$0.056}} & \textbf{0.209\textcolor{tvcg26_1st_revise}{$\pm$0.029}} & \textbf{0.938\textcolor{tvcg26_1st_revise}{$\pm$0.008}} & \textbf{0.935\textcolor{tvcg26_1st_revise}{$\pm$0.013}} & \textbf{0.134\textcolor{tvcg26_1st_revise}{$\pm$0.018}} \\

\midrule
\multirow{5}{*}{Scribble}
 & \multirow{1}{*}{PaintsTorch~\cite{Hati2019PaintsTorchAU}} &   0.075\textcolor{tvcg26_1st_revise}{$\pm$0.014} & 10.957\textcolor{tvcg26_1st_revise}{$\pm$0.37} & 0.504\textcolor{tvcg26_1st_revise}{$\pm$0.045} & 0.434\textcolor{tvcg26_1st_revise}{$\pm$0.014} & 0.866\textcolor{tvcg26_1st_revise}{$\pm$0.010} & 0.887\textcolor{tvcg26_1st_revise}{$\pm$0.010} & 0.363\textcolor{tvcg26_1st_revise}{$\pm$0.021}  \\
& \multirow{1}{*}{Diffusart~\cite{CarrilloCVPRW2023}}  &   0.045\textcolor{tvcg26_1st_revise}{$\pm$0.008} & 13.069\textcolor{tvcg26_1st_revise}{$\pm$0.37} & \underline{0.596\textcolor{tvcg26_1st_revise}{$\pm$0.058}} & 0.313\textcolor{tvcg26_1st_revise}{$\pm$0.032} & 0.903\textcolor{tvcg26_1st_revise}{$\pm$0.012} & 0.911\textcolor{tvcg26_1st_revise}{$\pm$0.013} & 0.256\textcolor{tvcg26_1st_revise}{$\pm$0.025} \\
   & \textcolor{tvcg26_revise}{\multirow{1}{*}{ColDiffv1 \textcolor{tvcg26_1st_revise}{(200ep)}~\cite{Yan_2025_WACV}}} &  \textcolor{tvcg26_revise}{0.053\textcolor{tvcg26_1st_revise}{$\pm$0.008}} & \textcolor{tvcg26_revise}{13.698\textcolor{tvcg26_1st_revise}{$\pm$0.77}}  & \textcolor{tvcg26_revise}{0.510\textcolor{tvcg26_1st_revise}{$\pm$0.063}}  & \textcolor{tvcg26_revise}{0.372\textcolor{tvcg26_1st_revise}{$\pm$0.052}}
    & \textcolor{tvcg26_revise}{0.871\textcolor{tvcg26_1st_revise}{$\pm$0.037}} & \textcolor{tvcg26_revise}{0.899\textcolor{tvcg26_1st_revise}{$\pm$0.034}} & \textcolor{tvcg26_revise}{0.346\textcolor{tvcg26_1st_revise}{$\pm$0.093}}\\
   & \textcolor{tvcg26_revise}{\multirow{1}{*}{ColDiffv2 \textcolor{tvcg26_1st_revise}{(200ep)}~\cite{yan2025colorizediffusionv2enhancingreferencebased}}} &  \textcolor{tvcg26_revise}{\underline{0.041\textcolor{tvcg26_1st_revise}{$\pm$0.004}}} & \textcolor{tvcg26_revise}{\underline{14.925\textcolor{tvcg26_1st_revise}{$\pm$0.50}}}  & \textcolor{tvcg26_revise}{0.585\textcolor{tvcg26_1st_revise}{$\pm$0.034}}  & \textcolor{tvcg26_revise}{\underline{0.298\textcolor{tvcg26_1st_revise}{$\pm$0.020}}}
    & \textcolor{tvcg26_revise}{\underline{0.917\textcolor{tvcg26_1st_revise}{$\pm$0.012}}} & \textcolor{tvcg26_revise}{\underline{0.928\textcolor{tvcg26_1st_revise}{$\pm$0.016}}} & \textcolor{tvcg26_revise}{\underline{0.208\textcolor{tvcg26_1st_revise}{$\pm$0.030}}}\\
     & \multirow{1}{*}{DetFill (Ours)}&   \textbf{\textcolor{tvcg26_1st_revise}{0.015$\pm$0.005}} & \textbf{\textcolor{tvcg26_1st_revise}{19.723$\pm$1.41}} & \textbf{\textcolor{tvcg26_1st_revise}{0.724$\pm$0.086}} & \textbf{\textcolor{tvcg26_1st_revise}{0.183$\pm$0.047}} & \textbf{\textcolor{tvcg26_1st_revise}{0.954$\pm$0.010}} & \textbf{\textcolor{tvcg26_1st_revise}{0.951$\pm$0.011}} & \textbf{\textcolor{tvcg26_1st_revise}{0.084$\pm$0.021}}  \\
    \midrule
  \end{tabular}
\end{table*}

\begin{table}[t]
   \caption{\textbf{\tvcgrevii{Hint-AUC on Danbooru2021 with a fixed random region order.}}
   \tvcgrevii{Each image uses one area-independent order established during dataset generation; no per-run hint-selection seed is used. Values are the mean $\pm$ sample standard deviation across the three sketch sources, not across random seeds. See Sec.~VI-A for the inference-seed policy and Table~\ref{tbl:exp1-result-hintauc-deterministic} for the definition of \emph{200ep}.}
   }%
  \label{tbl:exp1-result-hintauc-random-sampling-simple}
  \centering
  \footnotesize
  \setlength{\tabcolsep}{3pt}
  \resizebox{\columnwidth}{!}{%
  \begin{tabular}{cc ccc}
    \toprule
     \multirow{2}{*}{Hint} & \multirow{2}{*}{Model} &   \multicolumn{3}{c}{Hint-AUC} \\
     & &  MSE ($\downarrow$) &  LPIPS ($\downarrow$) &  DreamSim ($\downarrow$)   \\
\midrule
\multirow{5}{*}{Dot}
& \multirow{1}{*}{PaintsTorch} & \underline{0.083\textcolor{tvcg26_1st_revise}{$\pm$0.002}} &  0.486\textcolor{tvcg26_1st_revise}{$\pm$0.024} &  0.467\textcolor{tvcg26_1st_revise}{$\pm$0.018}  \\
& \multirow{1}{*}{Diffusart}  &   0.101\textcolor{tvcg26_1st_revise}{$\pm$0.004} & 0.409\textcolor{tvcg26_1st_revise}{$\pm$0.008} & 0.331\textcolor{tvcg26_1st_revise}{$\pm$0.002} \\
& \textcolor{tvcg26_revise}{\multirow{1}{*}{ColDiffv1 \textcolor{tvcg26_1st_revise}{(200ep)}~\cite{Yan_2025_WACV}}} &  \textcolor{tvcg26_revise}{0.154\textcolor{tvcg26_1st_revise}{$\pm$0.006}} & \textcolor{tvcg26_revise}{0.697\textcolor{tvcg26_1st_revise}{$\pm$0.010}} & \textcolor{tvcg26_revise}{0.652\textcolor{tvcg26_1st_revise}{$\pm$0.031}}  \\
& \textcolor{tvcg26_revise}{\multirow{1}{*}{ColDiffv2 \textcolor{tvcg26_1st_revise}{(200ep)}~\cite{yan2025colorizediffusionv2enhancingreferencebased}}} &  \textcolor{tvcg26_revise}{0.084\textcolor{tvcg26_1st_revise}{$\pm$0.021}} & \textcolor{tvcg26_revise}{\underline{0.388\textcolor{tvcg26_1st_revise}{$\pm$0.059}}} & \textcolor{tvcg26_revise}{\underline{0.308\textcolor{tvcg26_1st_revise}{$\pm$0.054}}}  \\
& \multirow{1}{*}{DetFill (Ours)} &   \textbf{0.025\textcolor{tvcg26_1st_revise}{$\pm$0.005}} & \textbf{0.220\textcolor{tvcg26_1st_revise}{$\pm$0.031}} &  \textbf{0.139\textcolor{tvcg26_1st_revise}{$\pm$0.020}}  \\

\midrule
\multirow{5}{*}{Scribble}
& \multirow{1}{*}{PaintsTorch}  &  0.077\textcolor{tvcg26_1st_revise}{$\pm$0.013} &  0.435\textcolor{tvcg26_1st_revise}{$\pm$0.011} & 0.367\textcolor{tvcg26_1st_revise}{$\pm$0.019} \\
& \multirow{1}{*}{Diffusart}  &  \underline{0.053\textcolor{tvcg26_1st_revise}{$\pm$0.010}} &  \underline{0.322\textcolor{tvcg26_1st_revise}{$\pm$0.035}} & 0.257\textcolor{tvcg26_1st_revise}{$\pm$0.027}  \\
& \textcolor{tvcg26_revise}{\multirow{1}{*}{ColDiffv1 \textcolor{tvcg26_1st_revise}{(200ep)}~\cite{Yan_2025_WACV}}} &  \textcolor{tvcg26_revise}{0.157\textcolor{tvcg26_1st_revise}{$\pm$0.004}} & \textcolor{tvcg26_revise}{0.691\textcolor{tvcg26_1st_revise}{$\pm$0.014}} & \textcolor{tvcg26_revise}{0.662\textcolor{tvcg26_1st_revise}{$\pm$0.033}}  \\
& \textcolor{tvcg26_revise}{\multirow{1}{*}{ColDiffv2 \textcolor{tvcg26_1st_revise}{(200ep)}~\cite{yan2025colorizediffusionv2enhancingreferencebased}}} &  \textcolor{tvcg26_revise}{0.076\textcolor{tvcg26_1st_revise}{$\pm$0.006}} & \textcolor{tvcg26_revise}{0.338\textcolor{tvcg26_1st_revise}{$\pm$0.023}} & \textcolor{tvcg26_revise}{\underline{0.251\textcolor{tvcg26_1st_revise}{$\pm$0.036}}}  \\
& \multirow{1}{*}{DetFill (Ours)} &  \textbf{\textcolor{tvcg26_1st_revise}{0.021$\pm$0.007}} &  \textbf{\textcolor{tvcg26_1st_revise}{0.194$\pm$0.045}} & \textbf{\textcolor{tvcg26_1st_revise}{0.091$\pm$0.020}} \\
    \midrule
  \end{tabular}%
  }
\end{table}
\begin{figure*}[ht]
\centering
\begin{tabular}{c@{\hspace{0.01\linewidth}}c@{\hspace{0.01\linewidth}}c@{\hspace{0.01\linewidth}}c@{\hspace{0.01\linewidth}}c@{\hspace{0.01\linewidth}}c@{\hspace{0.01\linewidth}}}

\includegraphics[width = .11\linewidth]{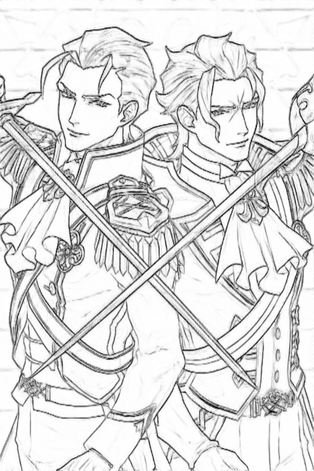} &
\includegraphics[width = .11\linewidth]{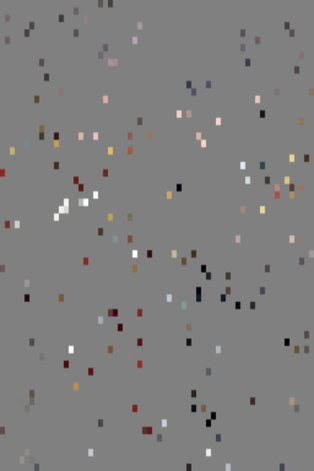} &
\includegraphics[width = .11\linewidth]{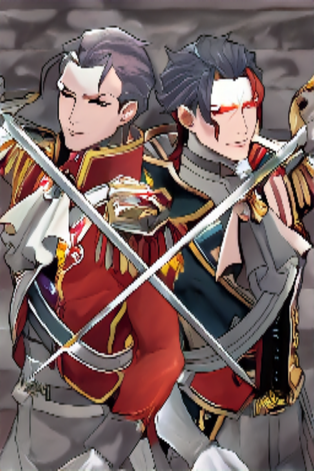} &
\includegraphics[width = .11\linewidth]{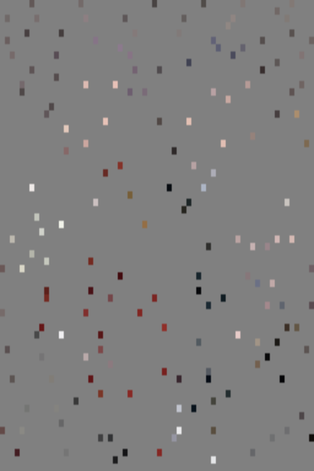} &
\includegraphics[width = .11\linewidth]{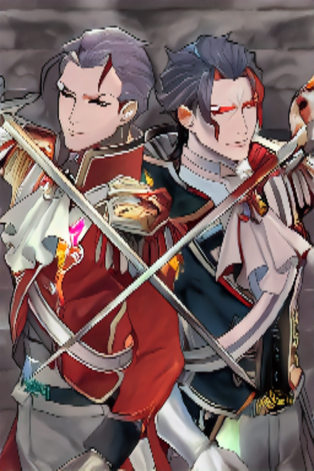} &
\hspace{0.5pt}\vrule\hspace{0.5pt}
\includegraphics[width = .11\linewidth]{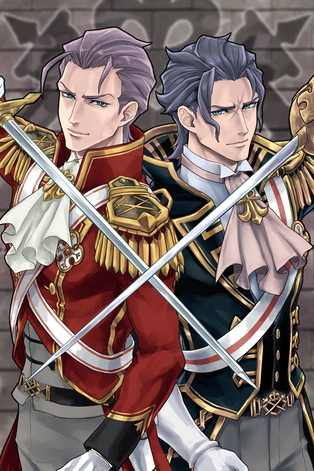} \\

\includegraphics[width = .11\linewidth]{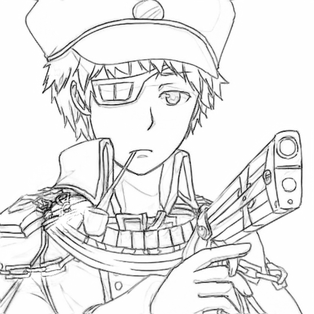} &
\includegraphics[width = .11\linewidth]{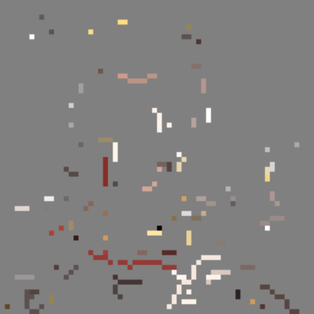} &
\includegraphics[width = .11\linewidth]{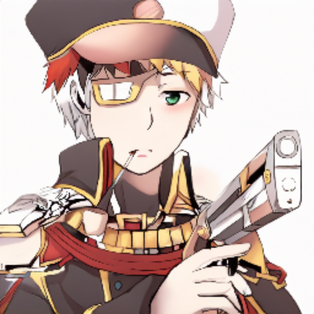} &
\includegraphics[width = .11\linewidth]{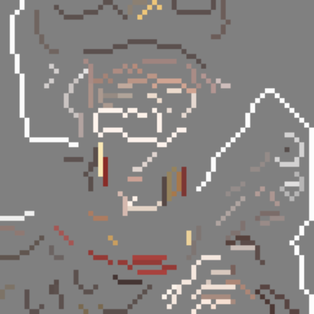} &
\includegraphics[width = .11\linewidth]{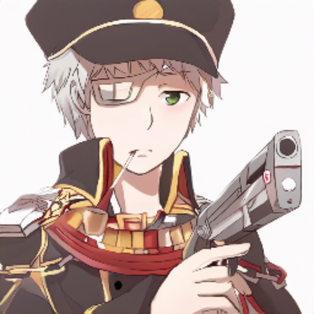} &
\hspace{0.5pt}\vrule\hspace{0.5pt}
\includegraphics[width = .11\linewidth]{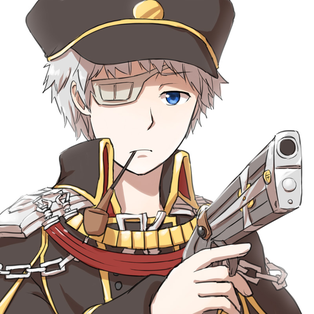} \\

{Input Sketch } & {Random Sample} & {Colorization} & {Region-based Sample} & {Colorization} &{GT} \\
 & {10 \%} & {10 \%} &{Top 10 \%}  & {Top 10 \%} & \\

\end{tabular}
   \caption{\textbf{Importance of deterministic hint sampling. }%
   We compare deterministic and random region-based hint sampling\tvcgrev{, with dot hints (top row) and scribble hints (bottom row)}.
   \textcolor{sig25_revise}{Top 10\% in region-based sample refers to the top 10\% of regions sorted according to pixel area.}
   Even with the same amount of regions being colorized, our deterministic approach follows a coarse-to-fine pattern, similar to artists, giving more consistent colorizations.
   }%
\label{fig:exp-visualization-different-hint-sampling-way}
\end{figure*}

\begin{table}[t]
   \caption{\textbf{\textcolor{tvcg26_revise}{A/B test score of the user study which is computed only from selection outcomes.}}
   \textcolor{tvcg26_revise}{The table presents the percentage of preferred scores, with a higher percentage indicating that users more frequently favored a method based on colorization quality.}
   \textcolor{tvcg26_1st_revise}{The user study was conducted with the models trained under the 200-epoch schedule (200ep).}
   }%
  \label{tbl:userstudy-small}
  \centering
  \begin{tabular}{c cccc}
    \toprule

       & PaintsTorch & Diffusart & \textcolor{tvcg26_revise}{ColDiffv2} & DetFill \\
       & \cite{Hati2019PaintsTorchAU} & \cite{CarrilloCVPRW2023} & \cite{yan2025colorizediffusionv2enhancingreferencebased} &  \\
       \midrule
       PaintsTorch~\cite{Hati2019PaintsTorchAU} & - & 20.8\%  & 24.8\% & 15.0\% \\
       Diffusart~\cite{CarrilloCVPRW2023} & 79.2\% & -  & 44.9\% & 32.6\% \\
       \textcolor{tvcg26_revise}{ColDiffv2}~\cite{yan2025colorizediffusionv2enhancingreferencebased}  & 75.2\% & 55.1\% & - &  32.2\%\\
       DetFill & 85.0\% & 67.4\% &  67.8\% & - \\      
    \midrule
  \end{tabular}
\end{table}

\textcolor{sig25_revise}{\section{Colorization Evaluation}}
\textcolor{tvcg26_revise}{
We perform experiments using our proposed evaluation framework
and compare our model with representative hint-based line art colorization methods from two major families:
(i) PaintsTorch~\cite{Hati2019PaintsTorchAU}, a GAN-based approach, and
(ii) diffusion-based approaches including Diffusart~\cite{CarrilloCVPRW2023}, \tvcgrev{ColorizeDiffusion}~v1 (ColDiffv1)~\cite{Yan_2025_WACV}, and \tvcgrev{ColorizeDiffusion}~v2 (ColDiffv2)~\cite{yan2025colorizediffusionv2enhancingreferencebased}.
}

\subsection{Implementation details}
\label{experiment-implement-details}
For training, we use the Adam optimizer with 200 epochs, a batch size of 20, and 5 NVIDIA RTX 2080Ti GPUs.
Due to memory constraints, we use gradient accumulation for both
Diffusart~\cite{CarrilloCVPRW2023} and our proposed DetFill.

We follow the training procedure of all approaches as closely as possible to the reference materials.
In particular, for PaintsTorch, we set a learning rate of $5e^{-5}$ with hyperparameters $\beta_{1}=0.5$, $\beta_{2}=0.9$.
For Diffusart and DetFill, we use a learning rate of $2e^{-5}$ and $1e^{-4}$, respectively.
All line art and color scribble images are resized to $256\times 256$, and values are normalized to the range $[-1, 1]$.
The number of time steps of DetFill is set to 1000 during training and 200 during inference.
\tvcgrev{Details of the checkpoint configurations and architecture are
provided in the supplementary material.}

\noindent\textbf{Synthetic Line Art.}
We create synthetic line arts from Danbooru2021~\cite{danbooru2021} using three extraction methods: XDoG~\cite{winnemoller2012xdog},
SketchKeras~\cite{sketchkeras}, and sketch simplification~\cite{simo2018mastering}.
The type of sketch is randomly sampled uniformly from the three types, with equal probability for each.

\noindent\textbf{Deterministic Color Hints.}
As the different approaches use different hint sizes, we use 64 $\times$ 64 px hints for PaintsTorch~\cite{Hati2019PaintsTorchAU}, and 256 $\times$ 256 px for Diffusart~\cite{CarrilloCVPRW2023} and DetFill.
For a fair comparison among colorization approaches, we generate the 64 $\times$ 64 px deterministic hints and then upsample \tvcgrev{(nearest neighbor)} the input size to 256 $\times$ 256 when necessary.
To generate the hints, we first perform segmentation at the original resolution and then downsample the results to 64 $\times$ 64 px when sampling.

For PaintsTorch~\cite{Hati2019PaintsTorchAU}, we use the author-provided random hint sampling algorithm directly.
For Diffusart~\cite{CarrilloCVPRW2023}, the hints are randomly sampled by averaging the colors of individual hints to determine a single uniform color, with the number of sampled hints set between $(4, 25)$, as specified in the original paper.
\textcolor{tvcg26_revise}{
We adapt \tvcgrev{ColorizeDiffusion}~v1~\cite{Yan_2025_WACV} and v2~\cite{yan2025colorizediffusionv2enhancingreferencebased} 
to the hint-based colorization setting 
by retraining both models using our deterministic hints as inputs.
Since these methods were originally formulated for reference-image conditioning rather than random hint sampling, 
using deterministic hints \tvcgrev{supports} fair comparisons of model performance.
}
For our approach, we use our proposed deterministic hints for training.
\tvcgrevii{We distinguish fixed hint inputs from stochastic inference. For each image, hint ratio, and sketch source, we evaluate one output per model. DetFill uses a fixed seed (1234) for all random-number generators of the implementation. We use the official inference scripts for the baselines: PaintsTorch is deterministic, Diffusart uses a fixed sampler seed, and ColorizeDiffusion v1/v2 use unseeded noise.}

\textcolor{tvcg26_1st_revise}{Both ColorizeDiffusion models start from the
officially released checkpoints and are fine-tuned with the common 200-epoch
training protocol used for all baselines above (denoted \emph{200ep}), thereby
retaining the large-scale pretraining of the original models.}

\noindent\textbf{Evaluation Metrics.}
For computing Hint-AUC, we use fixed hint ratios $\alpha \in \{ 0.00, 0.01, 0.03, 0.05, 0.10, 0.25, 0.50, 1.00 \}$.
As metrics, we consider MSE, PSNR, SSIM, LPIPS, OpenCLIP, DINO, and DreamSim.
We use our proposed test split with deterministic images for evaluation and
consider three settings corresponding to the different line art pre-processing
methods~\cite{winnemoller2012xdog, sketchkeras, simo2018mastering}.
Specifically, we use 20k images for training, 3k images for validation, and 3k images for testing.

\subsection{Comparisons with Existing Approaches}

Fig.~\ref{fig:exp1_visualization_colorization} summarizes the colorization results for different hint sampling ratios.
We observe that the performance of the \textcolor{tvcg26_1st_revise}{compared baseline methods} does not show a clear correlation with the hint sampling ratios.
In contrast, the proposed DetFill method consistently improves colorization results as the hint sampling ratio increases,
\textcolor{sig25_revise}{indicating that it is easier to refine and control the final colorization with hints using our approach.}

Table~\ref{tbl:exp1-result-hintauc-deterministic} shows a summary of Hint-AUC by region-based deterministic hints.
Our proposed DetFill shows overall better performance with both dot and scribble in all metrics.
\tvcgrevii{Table~\ref{tbl:exp1-result-hintauc-random-sampling-simple} uses ascending region-label order rather than area order. Labels were assigned from a palette shuffled once during dataset generation, giving each image one fixed, area-independent order. No hint-selection seed varies across runs. Supplementary Sec.~IV reports a separate DetFill experiment with ten hint-selection seeds.}
\tvcglayout{DetFill ranks first on all reported metrics in Tables~\ref{tbl:exp1-result-hintauc-deterministic} and~\ref{tbl:exp1-result-hintauc-random-sampling-simple}.}
\tvcgrevii{The standard deviations measure variation across sketch sources, not random seeds.}
\tvcglayout{The supplementary material examines sensitivity to hint-selection seeds, the $\alpha$ grid, and the region segmenter.}
These findings suggest that region-based hint sampling during the training phase is important for developing a more accurate and intuitive colorization model.
\textcolor{tvcg26_1st_revise}{Retraining Diffusart with our deterministic
region-based hints under the same 200-epoch protocol also lifts it close to
DetFill (e.g., scribble DreamSim $0.084$ for both; Table~X in the
supplementary material), indicating that the gain comes from the
hint-generation and training protocol rather than a specific architecture.
The supplementary material additionally evaluates ColorizeDiffusion under the
authors' original 7-epoch fine-tuning schedule.}
\subsection{Importance of Region-based Sorting for Hint Sampling}

Fig.~\ref{fig:exp-visualization-different-hint-sampling-way} shows the results of the different hint sampling algorithms with the same sampling ratio.
With randomly sampled deterministic hints, the important parts of the color such as background, body, and light color are not well reflected in colorization results.
\tvcgrev{Region-based sorting improves the results for both hint types (dot
hints in the top row and scribble hints in the bottom row of
Fig.~\ref{fig:exp-visualization-different-hint-sampling-way}), and the effect}
is especially notable for scribble hints: the colorization results differ significantly, as can be clearly observed in the corresponding figures.
In the case of the deterministic sampling approach, the important color information is well reflected in the colorization results.
These results seem to indicate that the deterministic sampling plays an important role \textcolor{sig25_revise}{in} the Hint-AUC metric.

\vspace{0.15em}
\subsection{Colorization on the Natural Images}
Fig.~\ref{fig:supp_visualization_deterministic_scribble_real-simple}
visualizes the colorization results on natural images.
We replace the Danbooru2021 dataset with ImageNet~\cite{russakovsky2015imagenet} and train under
the same conditions as described in Section~\ref{experiment-implement-details}.
\textcolor{sig25_revise}{DetFill demonstrates promising performance on \tvcgrev{natural-image} examples, 
suggesting potential applicability beyond synthetic test cases. 
However, with fewer hints, the proposed approach still produces inconsistent colorizations
compared to the ground-truth image.}

\begin{table*}[t]
  \caption{\textbf{User study score per hint ratio over the deterministic hints with illustration dataset (Danbooru2021)}: \tvcgrev{User preferences vary substantially with the hint ratio $\alpha$ across models, supporting evaluation at multiple hint ratios.} \textcolor{tvcg26_1st_revise}{The user study was conducted with the models trained under the 200-epoch schedule (200ep).}}
  \label{tbl:userstudy-small-ratiowise}
  \centering
  \begin{tabular}{cc cccccccc}
    \toprule
       \multirow{2}{*}{Selected Model} & \multirow{2}{*}{Compared Model} &   \multicolumn{8}{c}{Hint Ratio (\%) } \\
      & &    0\% & 1\% & 3\%  & 5\% & 10\% & 25\%  & 50\% & 100\%  \\
      \midrule
      DetFill & PaintsTorch~\cite{Hati2019PaintsTorchAU} &  33.6\% & 83.6\% & 85.9\% & \tvcgrevii{89.8\%} & 96.9\% & 94.5\% & 96.1\% & 99.2\% \\
      DetFill & Diffusart~\cite{CarrilloCVPRW2023} &  19.5\% & 54.7\% & 53.1\% & 78.9\% & 71.1\% & 79.7\% & 90.6\% & 91.4\% \\
      DetFill & \textcolor{tvcg26_revise}{ColDiffv2}~\cite{yan2025colorizediffusionv2enhancingreferencebased} & 64.8\% & 55.5\% & 68.0\% & 64.1\% & 64.1\% & 63.3\% & 85.9\% & 76.6\% \\

      \midrule
  \end{tabular}
\end{table*}

\providecommand{\qimg}[1]{}
\providecommand{\qimgm}[2]{%
  \raisebox{-.4\height}{\includegraphics[width=.09\linewidth]{#1}}%
}

\begin{figure*}[ht]
\centering
\setlength{\tabcolsep}{8pt} %
\renewcommand{\arraystretch}{1.0}
\scriptsize

\begin{tabular}{c c c c c c c c c}

\raisebox{4\height}{$1\%$} &
\qimgm{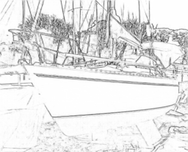}{sketch} &
\qimgm{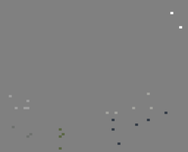}{hint} &
\qimgm{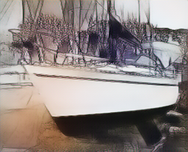}{PT} &
\qimgm{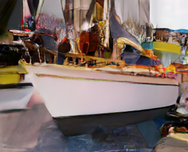}{Diffusart} &
\qimgm{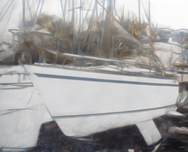}{Diffusart} &
\qimgm{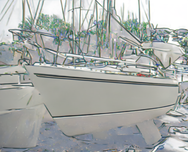}{Diffusart} &
\qimgm{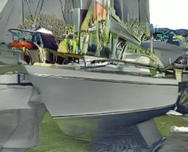}{DetFill} &
\qimgm{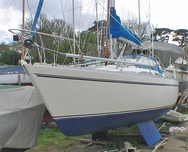}{GT} \\

\raisebox{4\height}{$10\%$} &
\qimgm{figs/imgs_sig25/supp/deterministic/real_dot/1610_sketch.png}{sketch} &
\qimgm{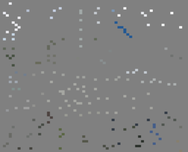}{hint} &
\qimgm{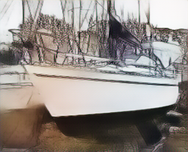}{PT} &
\qimgm{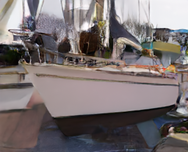}{Diffusart} &
\qimgm{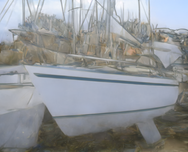}{Diffusart} &
\qimgm{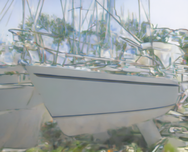}{Diffusart} &
\qimgm{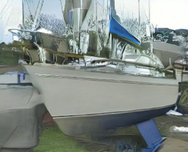}{DetFill} &
\qimgm{figs/imgs_sig25/supp/real_original/n02838178_849.png}{GT} \\

\raisebox{4\height}{$50\%$} &
\qimgm{figs/imgs_sig25/supp/deterministic/real_dot/1610_sketch.png}{sketch} &
\qimgm{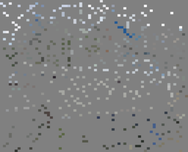}{hint} &
\qimgm{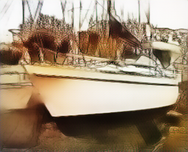}{PT} &
\qimgm{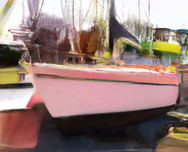}{Diffusart} &
\qimgm{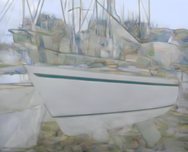}{Diffusart} &
\qimgm{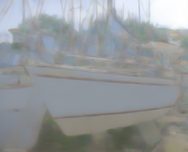}{Diffusart} &
\qimgm{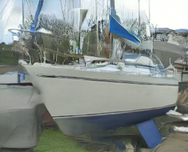}{DetFill} &
\qimgm{figs/imgs_sig25/supp/real_original/n02838178_849.png}{GT} \\

\raisebox{4\height}{$100\%$} &
\qimgm{figs/imgs_sig25/supp/deterministic/real_dot/1610_sketch.png}{sketch} &
\qimgm{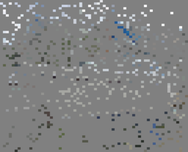}{hint} &
\qimgm{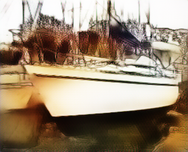}{PT} &
\qimgm{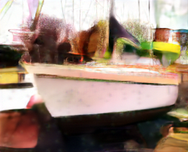}{Diffusart} &
\qimgm{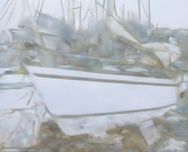}{Diffusart} &
\qimgm{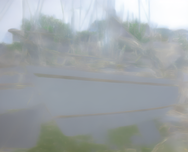}{Diffusart} &
\qimgm{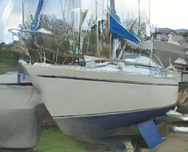}{DetFill} &
\qimgm{figs/imgs_sig25/supp/real_original/n02838178_849.png}{GT} \\

\raisebox{4\height}{$1\%$} &
\qimgm{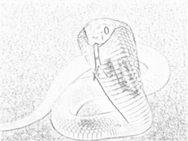}{sketch} &
\qimgm{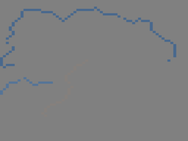}{hint} &
\qimgm{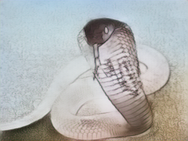}{PT} &
\qimgm{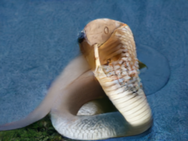}{Diffusart} &
\qimgm{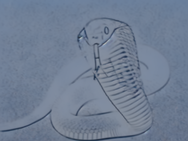}{Diffusart} &
\qimgm{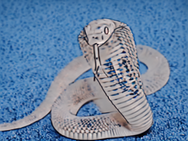}{Diffusart} &
\qimgm{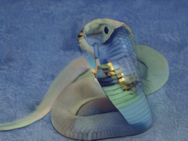}{DetFill} &
\qimgm{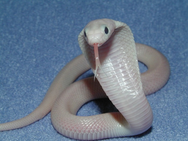}{GT} \\

\raisebox{4\height}{$10\%$} &
\qimgm{figs/imgs_sig25/supp/deterministic/real_scribble/hint/1/color_912_sketch.png}{sketch} &
\qimgm{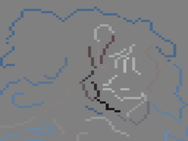}{hint} &
\qimgm{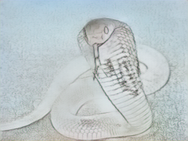}{PT} &
\qimgm{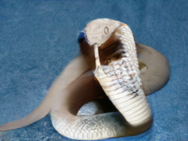}{Diffusart} &
\qimgm{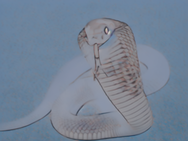}{Diffusart} &
\qimgm{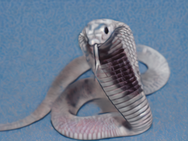}{Diffusart} &
\qimgm{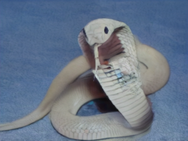}{DetFill} &
\qimgm{figs/imgs_sig25/supp/real_original/n01748264_6447.png}{GT} \\

\raisebox{4\height}{$50\%$} &
\qimgm{figs/imgs_sig25/supp/deterministic/real_scribble/hint/1/color_912_sketch.png}{sketch} &
\qimgm{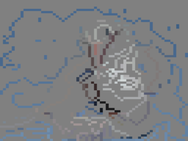}{hint} &
\qimgm{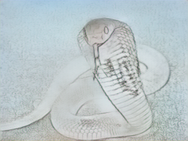}{PT} &
\qimgm{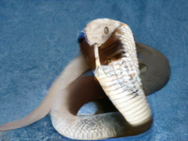}{Diffusart} &
\qimgm{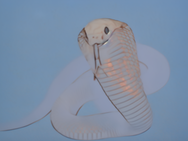}{Diffusart} &
\qimgm{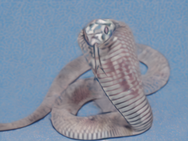}{Diffusart} &
\qimgm{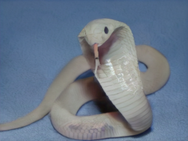}{DetFill} &
\qimgm{figs/imgs_sig25/supp/real_original/n01748264_6447.png}{GT} \\

\raisebox{4\height}{$100\%$} &
\qimgm{figs/imgs_sig25/supp/deterministic/real_scribble/hint/1/color_912_sketch.png}{sketch} &
\qimgm{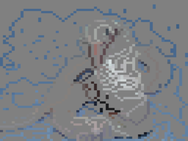}{hint} &
\qimgm{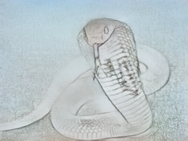}{PT} &
\qimgm{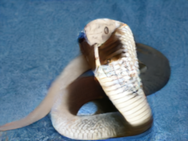}{Diffusart} &
\qimgm{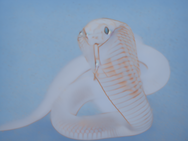}{Diffusart} &
\qimgm{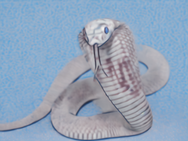}{Diffusart} &
\qimgm{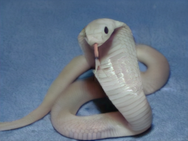}{DetFill} &
\qimgm{figs/imgs_sig25/supp/real_original/n01748264_6447.png}{GT} \\

\textbf{$\alpha$} &
\textbf{Sketch} &
\textbf{Hints} &
\shortstack{\textbf{PaintsTorch}} &
\shortstack{\textbf{Diffusart}} &
\shortstack{\textbf{ColDiffv1}} &
\shortstack{\textbf{ColDiffv2}} &
\textbf{DetFill} &
\textbf{GT} \\

\end{tabular}

\caption{\textbf{Colorization results on natural images from ImageNet~\cite{russakovsky2015imagenet}.}
Although the focus of our proposed approach is on the illustration domain,
\textcolor{sig25_revise}{our approach can perform colorization well on }\tvcgrev{natural
images and compares favorably with the baseline approaches.}
}
\label{fig:supp_visualization_deterministic_scribble_real-simple}
\end{figure*}

\section{Perceptual User Study}

To further validate the applicability of Hint-AUC, we conduct a user study using an A/B test with synthetic colorization results generated from ground-truth illustration images.
\textcolor{tvcg26_revise}{In particular, we prepared 192 test illustration images from Danbooru2021~\cite{danbooru2021} and generated synthetic images using PaintsTorch~\cite{Hati2019PaintsTorchAU}, Diffusart~\cite{CarrilloCVPRW2023}, \tvcgrev{ColorizeDiffusion}~v2 (ColDiffv2)~\cite{yan2025colorizediffusionv2enhancingreferencebased}, and the proposed model.
}
We tested hint ratios $\alpha \in \{ 0.00,\allowbreak 0.01,\allowbreak 0.03,\allowbreak 0.05,\allowbreak 0.10,\allowbreak  0.25,\allowbreak 0.50,\allowbreak 1.00 \}$.
To evaluate the results, we recruited 32 participants, resulting in 6,144 pairwise judgments in total (32 participants $\times$ 192 trials).
\tvcglayout{We used a within-subject design. Each participant completed 192 trials in randomized order: six method pairs, eight hint ratios, and four different images per condition.}
In the user study, two synthetic images and 
\textcolor{tvcg26_revise}{the corresponding sketch and hint images} were shown to the participants,
who were asked to select the image with higher perceptual quality given the sketch and hints.
\textcolor{tvcg26_revise}{The participants were fairly compensated for their time. All participants were university students or older with sufficient PC proficiency and provided informed consent via a signed consent form.}
\tvcgrev{This work involved human subjects; the complete ethics statement appears in the supplementary material.}

\textcolor{tvcg26_revise}{
As shown in Table~\ref{tbl:userstudy-small}, our method is preferred in 85.0\% of pairwise comparisons against PaintsTorch, 67.4\% against Diffusart, and 67.8\% against ColDiffv2.
}
\textcolor{tvcg26_revise}{Additionally, ColDiffv2 is preferred by 55.1\% over Diffusart, and Diffusart is preferred by \textcolor{tvcg26_1st_revise}{79.2\%} over PaintsTorch.}
\tvcgrev{These results are consistent with the quantitative evaluations in that DetFill is the top-ranked method in Tables~\ref{tbl:exp1-result-hintauc-deterministic} and~\ref{tbl:exp1-result-hintauc-random-sampling-simple}; the ordering among the baselines varies across metrics and sampling protocols.}
\textcolor{tvcg26_revise}{
We further break down the A/B preferences by hint ratio in Table~\ref{tbl:userstudy-small-ratiowise}.
We observe that the preference for DetFill changes with the hint ratio $\alpha$.
\tvcgrev{In particular, at $\alpha{=}1\%$, preferences are close to balanced against Diffusart and ColDiffv2 (54.7\% and 55.5\%) and remain close against Diffusart at $\alpha{=}3\%$ (53.1\%), while DetFill is clearly preferred over ColDiffv2 at $3\%$ (68.0\%) and at larger $\alpha$.}
This supports our motivation to evaluate colorization models across multiple hint ratios rather than at a single hint setting.
}

\textcolor{tvcg26_revise}{
To assess the statistical significance of the A/B preferences while accounting for repeated measurements,
we analyze the binary choices using a generalized linear mixed-effects model~\cite{breslow1993approximate} with a logit link and random intercepts for participants and images.
The model confirms that DetFill is selected significantly more often than chance ($z=30.18$, $p<10^{-16}$).
Moreover, when modeling the compared baseline method, the hint ratio $\alpha$ (treated as a categorical factor), and their interaction as fixed effects,
we observe a significant main effect of $\alpha$ ($\chi^2(7)=618.33$, $p=2.75\times10^{-129}$) and a significant baseline$\times\alpha$ interaction
($\chi^2(14)=324.04$, $p=1.12\times10^{-60}$), indicating that DetFill's relative advantage depends on the hint ratio and varies across baselines,
consistent with Table~\ref{tbl:userstudy-small-ratiowise}.
}

\section{Conclusion}

\textcolor{sig25_revise}{
\tvcgrevii{We introduced Hint-AUC, a protocol for evaluating colorization across hint ratios. The evaluation results are consistent with the overall human-preference results.}
We also showed that Hint-AUC can be used not only for evaluation but also for improving the colorization results of existing approaches.
Moreover, we demonstrated that our DetFill model achieves strong and consistent performance across a wide range of hint ratios under the proposed framework.
We will release our full source code to facilitate reproducibility, and we hope it will encourage other researchers to adopt our evaluation framework.
}
\textcolor{sig25_revise}{
\tvcgrevii{Future work will explore segmentation methods that better reflect human perception. We will also} develop more efficient diffusion architectures to reduce inference 
time and parameter counts while maintaining high quality even with sparse hints.
}

\textcolor{tvcg26_1st_revise}{\paragraph{Limitations and scope}
Hint-AUC targets hint-conditioned line-art colorization at the standard
$256\times256$ academic setting; extending it to the 2K--4K resolutions of
professional workflows is left to future work. Models not trained with hint
conditioning require retraining or fine-tuning to accept hint inputs,
as we demonstrate with ColorizeDiffusion, and video colorization models such
as LVCD~\cite{huang2024lvcd} are outside the current scope. The
region-averaged hint colors (Sec.~IV) are a simplification shared with
existing uniform-color protocols, likely more consequential for natural
images; lighting-aware hint generation remains future work.
\tvcgrevii{The protocol can use other deterministic region-to-stroke rules, but we evaluate only longest-path scribbles and dots. Models trained with other hint distributions may encounter a distribution shift under size-ordered evaluation, so we report both selection rules. The stored hint maps and selection rule define the evaluation inputs; inference seeds are reported in Sec.~VI-A. DetFill ranks first under both selection rules (Tables~\ref{tbl:exp1-result-hintauc-deterministic}--\ref{tbl:exp1-result-hintauc-random-sampling-simple}), consistent with the overall preference for DetFill in the user study (Tables~IV--V).}}

\bibliographystyle{IEEEtran}
\bibliography{egbib}

\vspace{-2.1\baselineskip}
\begin{IEEEbiographynophoto}{Koki Madono}
\tvcgrev{is with Waseda University, working on line art colorization.}
\end{IEEEbiographynophoto}

\vspace{-2.2\baselineskip}
\begin{IEEEbiographynophoto}{Yuan Mingcheng}
\tvcgrev{is with Waseda University, working on image colorization.}
\end{IEEEbiographynophoto}

\vspace{-2.4\baselineskip}
\begin{IEEEbiographynophoto}{Edgar Simo-Serra}
\tvcgrev{is with Waseda University, working on machine learning.}
\end{IEEEbiographynophoto}

\end{document}